\documentclass[twocolumn,final]{svjour3}          % twocolumn
\smartqed  % flush right qed marks, e.g. at end of proof
\usepackage{graphicx}
\usepackage{color}
\usepackage{amssymb}
\usepackage{booktabs}
\usepackage{amsmath}
\usepackage{multirow}
\usepackage[colorlinks,linkcolor=red,anchorcolor=blue,citecolor=blue,CJKbookmarks=True]{hyperref}
\usepackage{natbib}%\usepackage[numbers]{natbib} %When citations are numbered,
\usepackage[linesnumbered,ruled,vlined]{algorithm2e}
\begin{document}
%\thanks{Grants or other notes
%about the article that should go on the front page should be
%placed here. General acknowledgments should be placed at the end of the article.}
%
%\title{Domain-Supervised Multimodal Image-to-Image Translation}
%
\title{Unsupervised Multi-Domain Multimodal Image-to-Image Translation with Explicit Domain-Constrained Disentanglement}
%\title{Unsupervised Multi-Domain Multimodal Image-to-Image Translation with Domain-Supervision and Explicit Disentangled Representation}
\titlerunning{DCM$^2$IT}
%
%\subtitle{Do you have a subtitle?\\ If so, write it here}
%\titlerunning{Short form of title}        % if too long for running head

\author{Weihao XIA \and  Yujiu Yang \and Jing-Hao Xue}

\authorrunning{Weihao XIA et al.} % if too long for running head

\institute{Weihao XIA  \at
              Graduate School at Shenzhen, Tsinghua University, P.R. China \\
%            Tel.: +123-45-678910\\
              \email{xiawh16@mails.tsinghua.edu.cn}           %  \\
%             \emph{Present address:} of F. Author  %  if needed
           \and
           Yujiu Yang\at
             Graduate School at Shenzhen, Tsinghua University, P.R. China \\
             \email{yang.yujiu@sz.tsinghua.edu.cn}
           \and
           Jing-Hao Xue\at
            Department of Statistical Science, University College London \\
            \email{jinghao.xue@ucl.ac.uk}
}

\date{Received: date / Accepted: date}
% The correct dates will be entered by the editor

\maketitle

\begin{abstract}
Image-to-image translation has drawn great attention during the past few years. It aims to translate an image in one domain to a given reference image in another domain. Due to its effectiveness
and efficiency, many applications can be formulated as image-to-image translation problems. However, three main challenges remain in image-to-image translation: 1) the lack of large amounts of aligned training pairs for different tasks; 2) the ambiguity of multiple possible outputs from a single input image; and 3) the lack of simultaneous training of multiple datasets from different domains within a single network. We also found in experiments that the implicit disentanglement of content and style could lead to unexpect results. In this paper, we propose a unified framework for learning to generate diverse outputs using unpaired training data and allow simultaneous training of multiple datasets from different domains via a single network. Furthermore, we also investigate how to better extract domain supervision information so as to learn better disentangled representations and achieve better image translation. Experiments show that the proposed method outperforms or is comparable with the state-of-the-art methods.

% \PACS{PACS code1 \and PACS code2 \and more}
% \subclass{MSC code1 \and MSC code2 \and more}
\end{abstract}

\section{Introduction}
\label{intro}

\begin{table*}[t]
  \centering
 \caption{Feature-by-feature comparison of image-to-image translation methods. Our model achieves unsupervised multi-domain multi-modal image-to-image translation with explicit domain-constrained
disentanglement.}
\label{tab:feature-by-feature comparison}
%\resizebox{\textwidth}{12mm}
\setlength{\tabcolsep}{0.8mm}
{
\begin{tabular}{cccccccccc}
\toprule
Method                                          & Pix2pix   	  & CycleGAN 	& BicycleGAN	    & StarGAN 	     & DosGAN 	 	    & MUNIT	       & DRIT		       & Ours		\\
\midrule
Unsupervised learning                  &	-  	             & \checkmark	&	- 	                        & \checkmark	 & \checkmark		& \checkmark	& \checkmark	& \checkmark	\\
Multi-modal                                   &	-	             & - 		                & \checkmark        & - 		                 &- 				             & \checkmark	& \checkmark	& \checkmark	 \\
Multi-domain 	                                & 	-	         & -		                & -		                    & \checkmark	 &\checkmark	     & - 				        & - 		                & \checkmark\\
Disentangled representation         & 	-	             & -		                & -		                    & -	                     &-	                         & \checkmark	& \checkmark 	& \checkmark\\
Domain supervision                      & 	-	             & -		                & -		                    & -	                     &\checkmark	     & - 				        & - 		                & \checkmark\\
\bottomrule
\end{tabular}
}
\end{table*}

Image-to-image translation aims to learn a mapping that can transfer an image from a source domain to a target domain, while maintaining the main representations of the input image. It has received significant attention since many problems in computer vision can be formulated as the cross-domain image-to-image translation \citep{isola2017image,CycleGAN2017,zhu2017toward}, including super-resolution \citep{Ledig2016Photo}, image inpainting \citep{yu2018free,yu2018generative,nazeri2019edgeconnect}, and style transfer \citep{Gatys2016Image}.

Despite of the great success, learning the mapping between two visually different domains is still challenging in three aspects. First, exquisite large-scale datasets with thousands of aligned training pairs for different tasks are often unavailable. Second, in many scenarios, such mappings of interest are inherently multi-modal or one-to-many, i.e., a single input may correspond to multiple possible outputs. Third, for multi-domain image translation tasks, many existing methods learn an individual mapping separately between only two domains selected from all given domains. With $n$ domains, this needs $\tbinom{n}{2}=\Theta\left(n^{2}\right)$ models to train. They are incapable of jointly learning the mapping between all available domains from different datasets.  Several recent efforts have been made to address these issues.

To tackle the paired training data limitation, many works propose their unsupervised learning frameworks for general-purpose image-to-image translation. Most methods are inspired by the idea that the unpaired images from two domains should be consistent with their reconstructions in a cyclic mapping \citep{CycleGAN2017} or primal-dual relation \citep{yi2017dualgan}. Superiority of this cycle consistency loss has been demonstrated on several tasks where paired training data hardly exist. However, these methods fail to produce multi-modal outputs conditioned on the given input image.

To capture the full distribution of possible outputs, simply incorporating noise vectors as additional inputs often lead to the mode collapsing issue and thus does not increase the variation of the generation images. \citet{zhu2017toward} tries to encourage the one-to-one relationship between the output and the latent vector to generate diverse outputs. Nevertheless, the training process of \citet{zhu2017toward} requires paired images. Very recently, \citet{DRIT} and \citet{huang2018munit} propose the disentangled representation framework to generate diverse outputs with unpaired training data. These two multi-modal unsupervised image-to-image translation methods assume that the latent space of image can be decomposed into a content latent space and a style latent space, and the images in different domains vary in the style but share a common content. Thus multi-modality can be achieved by recombining the content vector of an image from the source domain with a random style vector in the target style latent space.

To simultaneously train multiple datasets with different domains within a single network, \citet{StarGAN2018} uses a label (e.g., binary or one-hot vector) to represent domain information. Instead of learning a fixed mapping for two domains, they input both images and their corresponding domain information to the model, and learn to flexibly translate the images from the source domain to the target domain. By controlling domain labels, an image can be translated into any desired domain. Instead of representing domain characteristics with multiple domain labels as in \citet{StarGAN2018}, \citet{lin2019unpaired} treats domain information as explicit supervision. They pre-train a classification network to classify the domain of an image. Such features, together with the latent content features of an image in the source domain are used to generate an image in the target domain. Such features extracted from this pre-trained network can represent domain information, thus can be called domain features and training with domain features is called Domain Supervision. However, both methods produce a single output and are lack of output diversity.

Many methods \citep{DRIT,huang2018munit,liu2018unified} adapt disentangled representations for unsupervised image-to-image translation, but we found in experiments that implicit disentanglement learning may confuse content with style in some cases. As shown in Figure \ref{fig:drit_deblur}, if adapting \citet{DRIT} for image de-blurring task, the de-blurred images have different face contour with original one, which means that the attribute extractor has not only learned blur distortion pattern but also recognize some content representations like face contour as style. It can be attributed to the ambiguous and implicit disentanglement of content and style.

What's more, domain information are under-exploited in the area of image-to-image translation. For photo-to-art translation, we can distinguish that the generated image is either in the style of Pablo Picasso or in the style of Isaac Levitan. Similarly, different weather like sunny, foggy, rainy, snowy and cloudy should contain specific modality, and the same is true for seasons. Style itself could be learned in the collections of a unique artist.

To the best of our knowledge, we are the first image-to-image translation approach trying to handle forementioned challenges. In this paper, we propose the unsupervised \textbf{M}ulti-domain \textbf{M}ultimodal \textbf{I}mage-to-image \textbf{T}ranslation with Explicit \textbf{D}omain-\textbf{C}onstrained Disentanglement (named DCM$^2$IT). DCM$^2$IT is a unified framework for learning to generate diverse outputs with unpaired training data and allow simultaneous training of multiple datasets with different domains by a single network. Furthermore, we investigate how to utilize domain information and explicitly constrain the disentanglement for unsupervised image-to-image translation.

%Furthermore, many methods adapt disentangled representations for unsupervised image-to-image translation, but we found in experiments that implicit disentanglement might lead to confusion of content and style in some cases. We investigate how to utilize domain information and explicitly constrain the disentanglement so as to achieve better image translation.

To sum up, our key contributions can be summarized as:
\begin{itemize}
\item We introduce the first image-to-image method that achieves diverse outputs with simultaneous unsupervised training of multiple datasets by a single network.
\item We propose disentanglement learning constraint with domain supervision. We investigate how to extract domain supervision information so as to learn explicit disentangled representations of content and style.
\item Extensive qualitative and quantitative experiments on multiple datasets show that the proposed method outperforms or comparable with the state-of-the-art methods.
\end{itemize}

\section{Related work}
\label{sec:related work}
We initially provide an overview of the recent advances with Generative Adversarial Networks, then introduce some existing image-to-image translation methods and disentangled representations. We also give brief introduction of style transfer and domain adaptation, which are two tasks closely related with image-to-image translation.
\textbf{\subsection{Generative Adversarial Network}}
\label{sec:gan}
The Generative Adversarial Network (GAN) \citep{Goodfellow2014Generative} framework has achieved excellent results in many tasks such as image super-resolution \citep{Ledig2016Photo}, and image inpainting \citep{yu2018free,yu2018generative,nazeri2019edgeconnect}. GANs usually consist of a generator $G$ and a discriminator $D$. The training procedure for GANs is a minimax game between $G$ and $D$. $D$ is trained to distinguish whether the input image as real or fake, and $G$ is trained to fool $D$ with generated samples. The ideal solution is the Nash equilibrium where $G$ and $D$ could not improve their cost unilaterally \citep{heusel2017gans}.

Various improvement has been proposed to handle challenges in GANs including model generalization and training stability. \citet{arjovsky2017wasserstein} and \citet{Gulrajani2017Improved} propose to minimize the Wasserstein distance between model and data distributions. \citet{berthelot2017began} try to optimize a lower bound of the Wasserstein distances between auto-encoder loss distributions on real and fake data distributions. \citet{mao2017least} propose a least square loss for the discriminator, which implicitly minimizes Pearson $\mathcal{X}^2$ divergence, leading to stable training, high image quality and considerable diversity.

\textbf{\subsection{Image to image translation}}
\label{sec:i2i}

\citet{isola2017image} propose the first general  image-to-image translation method  (pix2pix) based on conditional GANs. \citet{wang2018pix2pixHD} propose an HD version of pix2pix by utilizing a coarse-to-fine generator, several multi-scale discriminators, and a feature matching loss, which increase the resolution to $2048 \times 1024$. Since it is usually time-consuming and expensive to collect such an exquisite large-scale dataset with thousands of image pairs, many studies have also attempted to tackle the paired training data limitation. \citet{CycleGAN2017}, \citet{kim2017learning}, \citet{yi2017dualgan} and \citet{DBLP:conf/nips/LiuBK17} leverage cycle consistency to regularize the unsupervised training process. Many works aim to produce diverse outputs, including \citet{zhu2017toward}, \citet{DRIT} and \citet{huang2018munit}. Some other methods like \citet{StarGAN2018}, \citet{lin2019unpaired}, \citet{liu2018unified} and \citet{anoosheh2018combogan} are proposed to improve the scalability of unsupervised image-translation methods. Table \ref{tab:feature-by-feature comparison} shows a feature-by-feature comparison among some existing image-to-image translation methods.

%Though the results of \citet{StarGAN2018} and \citet{lin2019unpaired} suggest a shared model for different domains similar enough to each other may be beneficial to the learning process. It is not clear whether the extracted domain features are really domain specific.

\begin{figure*}[th]
\centering
\includegraphics[width=0.8 \textwidth]{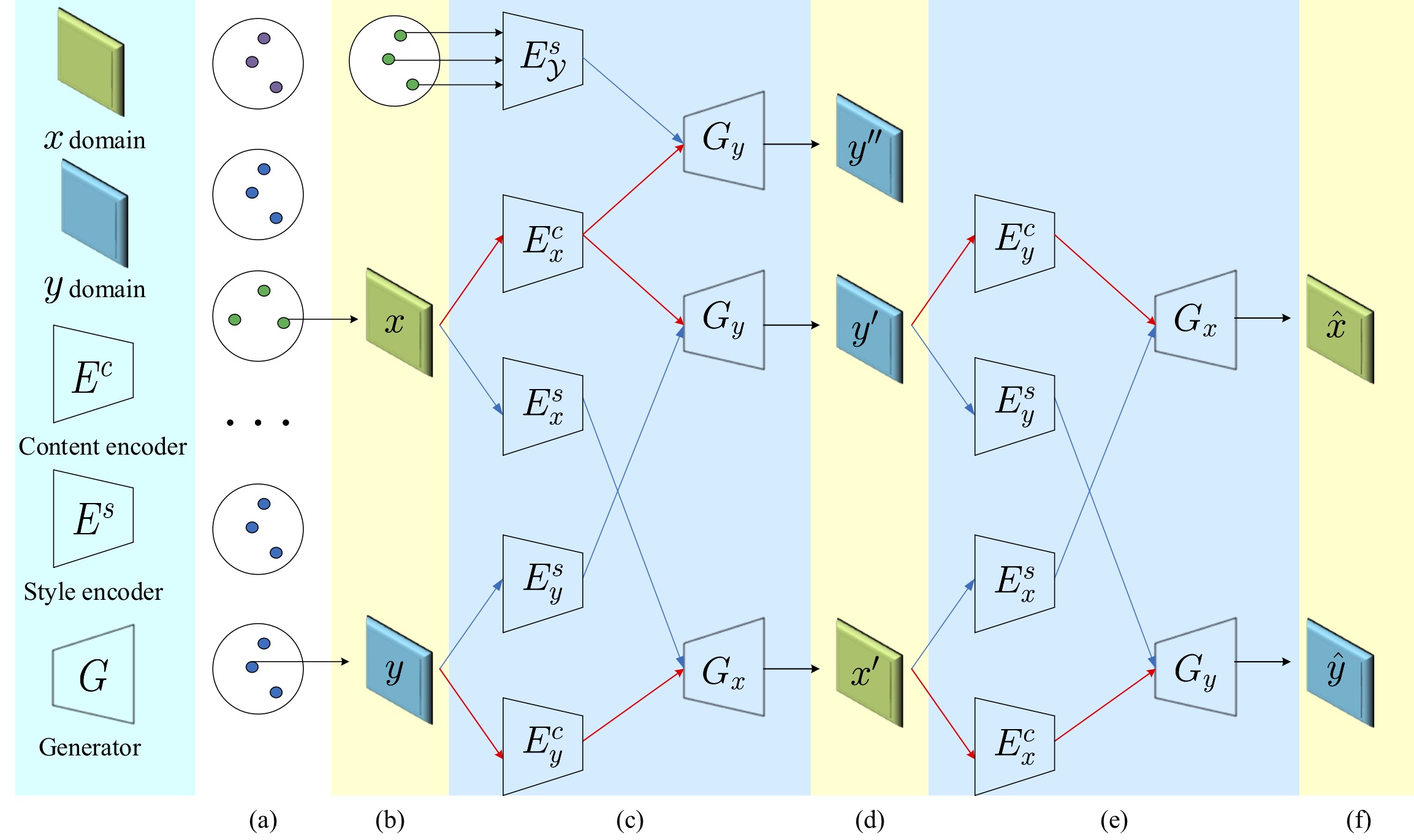}
\caption{The pipeline of our method. To achieve image translation between two domains, images from different domains are encoded as domain-invariant content representations $c$ and domain-specific style representations $s$. Then swap the style codes and use generator $G$ to produce the translated output images. The second translation process constrains the image reconstruction with cycle consistency loss. Due to the disentangled representations, the style representations are constrained to match the prior Gaussian distribution, so we can generate several possible outputs by random sampling from this prior. The domain style representations are extracted from collection of a certain style and constrain the disentanglement of content and style. The multi-domain simultaneous training is implemented by adding specific discriminative labels for each domain.}
\label{fig:pipeline}
\end{figure*}

\textbf{\subsection{Disentangled representations}}
\label{sec:disentangled}

There are many recent works on disentangled representation learning. For example, \citet{lu2019unsupervised} try to disentangle content from blur, \citet{denton2017unsupervised} separate time-independent and time-varying parts, \citet{Johnson2016Perceptual} intend to iteratively optimize the image by minimizing a content loss and a style loss, which can also be regard as an implicit disentanglement of content and style. \citet{zhu2017toward} combine cLR-GAN and cVAE-GAN to model the distribution of possible outputs. \citet{chen2016infogan} decompose representation by maximizing the mutual information between the latent factors and the synthesized images without utilizing paired training data. Some other works \citep{Xiao_2018_ECCV,liu2018unified,DRIT,huang2018munit} focus on disentanglement of content and style or attribute. It is difficult to explicitly define content or style and different works may adopt different definitions due to their specific tasks. In our setting, we refer to content as being visual elements that can be shared across domains and style as domain-specific. We disentangle an image into domain-invariant and domain-specific representations to facilitate learning diverse cross-domain mappings.

\textbf{\subsection{Style transfer}}
\citet{Gatys2016Image} introduce an impressive neural algorithm that transfers a content image to the style of another image, achieving so-called style transfer.  The original work of \citet{Gatys2016Image} iteratively updates the image to minimize a content loss and a style loss by a slow optimization process. Some methods \citep{Johnson2016Perceptual,li2016precomputed,ulyanov2016texture} change this optimization to a feed-forward alternative for acceleration. \citet{ulyanov2017improved} propose a method to improve the quality and diversity of the generated samples. \citet{chen2016fast} introduce a feed-forward method with style swap layers that can adapt to arbitrary styles even for those not observed during training.

Style transfer is closely related to image-to-image translation. However, image-to-image translation could not handle the tasks of example-guided style transfer, in which the target style is defined by a single image. When the target style is defined by a collection of images, image-to-image translation usually performs better than classical style transfer approaches.

\textbf{\subsection{Domain adaptation}}
\label{sec:domain adaption}

Domain adaptation is also similar with image-to-image translation. These approaches mainly focus on adapting a model trained in the source domain to another target domain.
The Adversarial Discriminative Domain Adaptation (ADDA) \citep{tzeng2017adversarial} aims to learn a discriminative feature subspace using the labeled source data. Then, it encodes the target data to this learned subspace using an asymmetric transformation through a domain-adversarial loss.
The Domain Adversarial Neural Network (DANN) \citep{bousmalis2016domain,ganin2016domain,tsai2018learning,ganin2014unsupervised} has focused on transferring deep representations by matching the feature distributions of different domains, aiming to learn domain-invariant features. In this case, it is critical to first define a measure of distance between two distributions. There are several different choices such as covariance \citep{sun2016deep}, Kullback-Leibler (KL) divergence \citep{kullback1951information}, and the non-parametric Maximum Mean Discrepancy (MMD) \citep{borgwardt2006integrating,long2014transfer,long2015learning,bousmalis2017unsupervised,zellinger2017central}.

\section{Method}
%\section{Method: Multi-Domain Multimodal Unsupervised Image-to-image Translation}

Our goal is to achieve unsupervised multi-domain multi-modal image-to-image translation via disentangled representations with a single model. The pipeline of our method is shown in Figure \ref{fig:pipeline}. For multi-domain translation, we design an intra-domain and inter-domain supervision mechanism, which is able to represent the essence of different domains and translate images cross different domains with only one single model. For multi-modal generation between two domains, we regularize the style codes in the training phase so that they can be represented by a Gaussian distribution. By controlling the parameters of style codes, multi-modality of generated images are possible. The model architecture and loss functions are also coherently designed for diverse and realistic image-to-image translation.
%The model architecture and loss functions are also coherently designed for diverse and realistic results using unpaired data.

\textbf{\subsection{Problem formulation}}

Assuming there are $n$ datasets of different domains $\{\mathcal{D}_{1},\mathcal{D}_{2},\cdots,\mathcal{D}_{n}\}$, our goal is to achieve unsupervised multi-domain multimodal image-to-image translation with domain-constrained disentanglement by single model. For each image $x_{i} \in \mathcal{D}_{i}$, the unique disentangled representations of content latent code $c \in \mathcal{C}$ and the style latent code $s_i \in \mathcal{S}_i$ can be extracted from content encoder $E_{\mathcal{D}_i}^{c}$ and style encoder $E_{\mathcal{D}_i}^{s}$. The generator $G_{i} $ can produce an image of certain style if given specific style latent code and corresponding content code.
Let $x_{1} \in \mathcal{D}_{1} $  and $x_{2} \in \mathcal{D}_{2}$  be images from two different domains, the content encoders $E_{1}^{c}$ and $E_{2}^{c}$  map images onto a domain-invariant content space $\left(E_{i}^{c} : \mathcal{D}_i \rightarrow \mathcal{C}\right)$ and the style encoders $E_{1}^{s}$  and $E_{2}^{s}$  map images onto the domain-specific style spaces $\left(E_{i}^{s} : \mathcal{D}_i \rightarrow \mathcal{S}_i\right)$. The generator  $G_{i}$ generates images conditioned on given content codes and style codes $\left( G_{i} :\left\{\mathcal{C}, \mathcal{S}_i \right\} \rightarrow \mathcal{D}_i\right)$.
We postulate that only the content latent part can be shared across domains and the style is part is domain-specific.

\textbf{\subsection{Intra-domain and inter-domain supervision}}
\label{sec:domain supervision}
\begin{figure*}[t]
\centering
  \includegraphics[width=0.8\textwidth]{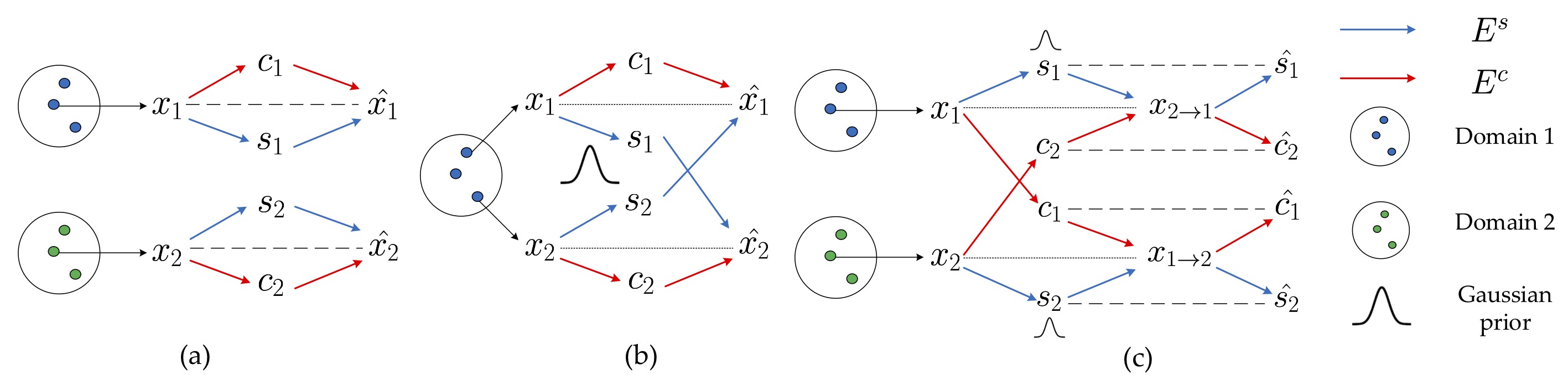}
\caption{Illustration of self translation, intra-domain translation and inter-domain translation. For better comprehension and comparison, we follow the representations in \citet{huang2018munit}, i.e. (a) and (b). To avoid confusion, we change their descriptions. Our model consists of two types of auto-encoders (denoted by  \textcolor{red}{red} and \textcolor{blue}{blue} arrows respectively), one for each domain. Similar with \citet{huang2018munit,DRIT}, the latent code of each auto-encoder is composed of a content code $c$ and a style code $s$. The model is trained with adversarial objectives (dotted lines) that ensure the translated images to be indistinguishable from real images in the target domain, as well as bidirectional reconstruction objectives (dashed lines) that reconstruct both images and latent codes.}
\label{fig:domain_tranlation}
\end{figure*}

To utilize domain information and explicitly constrain the disentanglement of content and style, we propose explicit domain-constrained disentanglement by first introducing intra-domain and inter-domain supervision.
%based on disentangled representation for unsupervised image-to-image translation.

Let $x_{1 \rightarrow 2}$  be a sample produced by translating image $x_1$  to its counterpart $x_2$  in domain $\mathcal{D}_{2}$ (similarly for  $x_{2 \rightarrow 1}$), then for a pair images $\left(x_{1}, x_{2}\right)$, we have
\begin{equation}
\begin{aligned}
x_1&=&G_1(c,s_1)&, &x_2&=&G_2(c,s_2), \\
%\end{aligned}
%\end{equation}
%\begin{equation}
%\begin{aligned}
x_{1 \rightarrow 2}&=&G_2(c,s_2)&,& x_{2 \rightarrow 1}&=&G_1(c,s_1).
\end{aligned}
\end{equation}

\begin{figure}[th]
\centering
  \includegraphics[width=0.25\textwidth]{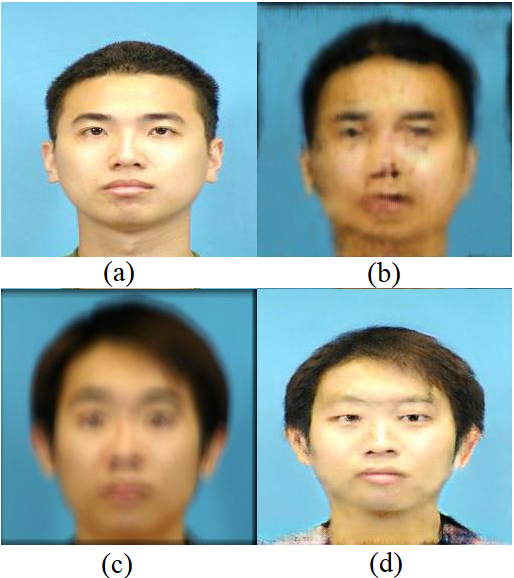}
\caption{Translation results of \citet{DRIT} for image de-blurring task. (a) real image. (b) blurred version of (a). (c) real blurred image. (d) deblurred version of (c). We can see that if adapting \citet{DRIT} for image de-blurring task, the de-blurred images have different face contours from original ones, which means that the attribute extractor has not only learned blur distortion pattern but also recognized some content representations such as face contour as attribute. It might be attributed to the ambiguous and implicit disentanglement of content and style.}
\label{fig:drit_deblur}
\end{figure}

Since $s_1$ and $s_2$ are domain-specific style codes extracted from single images $x_1$  and  $x_2$, respectively, we can call this translation as inter-domain translation. The style code extracted from a single image contains more information rather than only generalized style of a collection of images. In the training phase, the model may extracts incorrectly some content features as style features as illustrated in Figure \ref{fig:drit_deblur}.

To alleviate this situation, we design an intra-domain supervision to constrain the disentangled representation learning and represent the essence of different domains.
The main idea to achieve this is relatively simple: ``Two images from the same domain exchange their style codes, the generated images should be consistent with themselves.'' Different from style codes extracted from a single image $s_i \in \mathcal{S}_i$, these style codes extracted at domain level should be domain-specific and represent generalized domain style representations. For $n$ domains of datasets $\{\mathcal{D}_{1},\mathcal{D}_{2},\cdots,\mathcal{D}_{n}\}$, we have $n$ domains style representations $\{\mathcal{S}_{\mathcal{D}_{1}},\mathcal{S}_{\mathcal{D}_{2}},\cdots,\mathcal{S}_{\mathcal{D}_{n}}\}$. We can call this translation as intra-domain translation.
As shown in Figure \ref{fig:domain_tranlation}, intra-domain and inter-domain translation can be representation as

\begin{equation}
\begin{aligned}
&x_1&=&G_1(c,s_1), &x_2&=&G_2(c,s_2), \\
&x_{1 \rightarrow 2}&=&G_2(c,s_2),& x_{2 \rightarrow 1}&=&G_1(c,s_1), \\
&x_{1 \rightarrow 1'}&=&G_1(c,\mathcal{S}_{\mathcal{D}_1}),& x_{1' \rightarrow 1}&=&G_1(c,\mathcal{S}_{\mathcal{D}_1}).
\end{aligned}
\end{equation}

The intra-domain translation aims to learn the essence style of a domain, which means the learned style representations of images from the same domain do not vary to an unreasonable degree.  Specifically, all images converge to the ``mean'' style. After training on carefully selected images, this constraint helps the content and style encoders learn explicit disentangled representations during inter-domain translation. We can readily control its influence by changing the weight parameters.

\textbf{\subsection{Pre-training of domain style representation extractor}}

Different from many previous works regarding multiple domains as different sources of images, we treat each domain as explicit supervision. Similarly to \citet{lin2019unpaired}, we pre-train a domain feature representation extractor for each domain as explicit domain supervision.

For domain supervision, \citet{lin2019unpaired} train a classifier that tries to correctly distinguish images of different domains. Then they regard the output of second-to-last layer of the classifier as the domain feature. Different from this ambiguous and implicit definition, we try to learn the domain feature representations by intra-domain translation.

Given images from $n$ different domains, we train a CNN network by switching style codes of images from the same domain. The goal of this CNN network, which we call domain representation extractor $E_{\mathcal{D}_i }^{s}$, is to learn domain-specific style representations $\mathcal{S}_{\mathcal{D}_{1}}$ for domain $\mathcal{D}_i$  and to correctly classify the domain of an image. Then this pre-trained model  $E_{\mathcal{D}_i }^{s}$ is used as explicit domain supervision for inter-domain translation.

\begin{figure*}
  \includegraphics[width=1.0\textwidth]{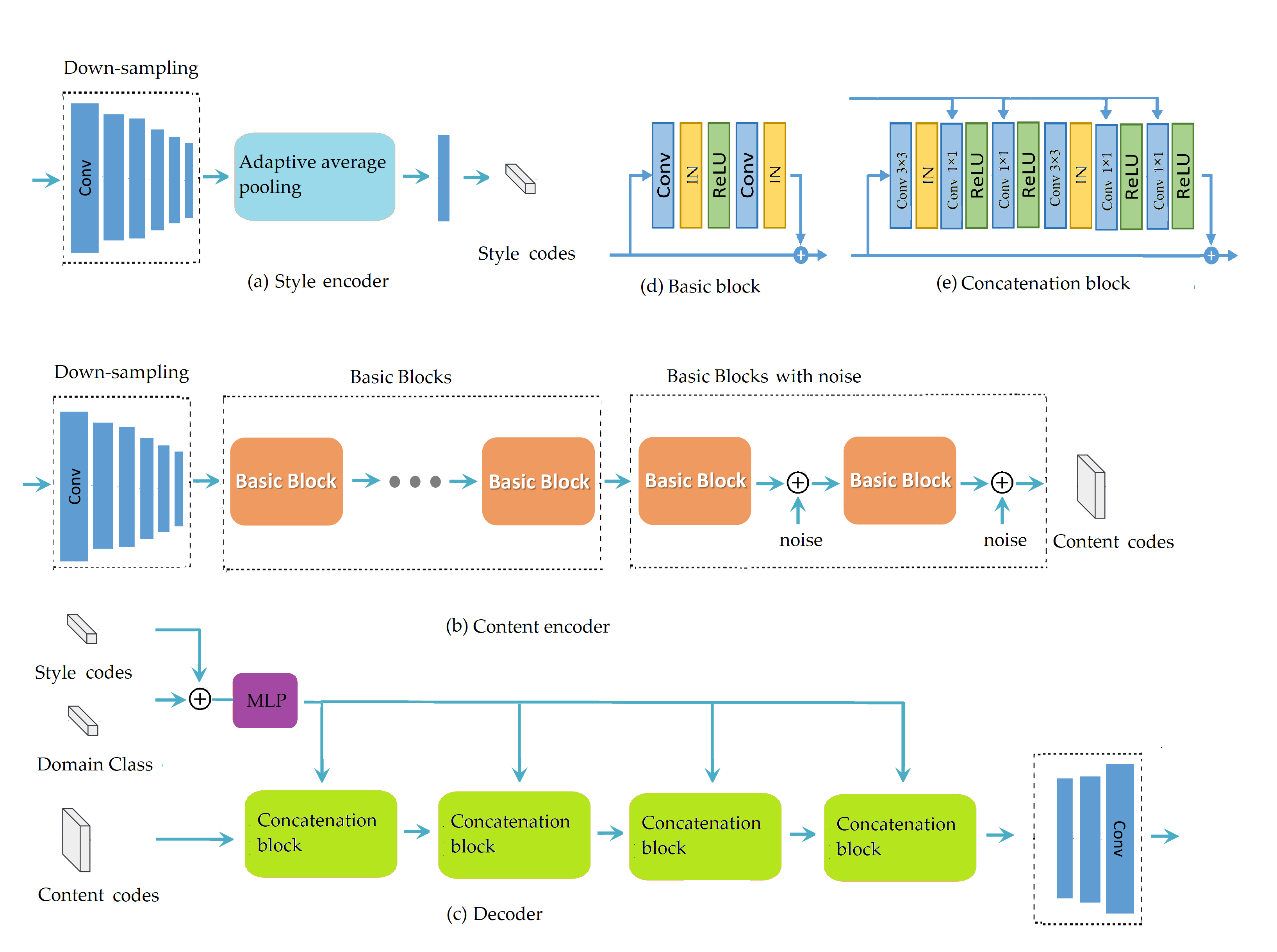}
\caption{Network architecture. For more details, refer to Section \ref{sec:network architecture}.}
\label{fig:network architecture}
\end{figure*}

\textbf{\subsection{Model}}
\label{sec:model}

As aforementioned, the pipeline of our model is shown in Figure \ref{fig:pipeline}. Similar to other unsupervised image-to-image translation via disentangled representations \citep{liu2018unified,DRIT,huang2018munit}, our model consists of a content encoder ${E}^c_i$ ,  style encoder ${E}^s_i $, decoder $G$  and discriminator $D_i$  for each domain $\mathcal{D}_i(i=1,2,\cdots,n)$. What's more, in our experiment, we have the domain classifier $D_{cls}$ pre-trained together with the domain style representation extractor $E_{\mathcal{D}_{i}}^{s}$.

As shown, to achieve image translation between two domains $\{\mathcal{D}_{1},\mathcal{D}_{2}\}$, images $x_1$, $x_2 $ from different domains are encoded as domain-invariant content representations $c_{1}=E_{1}^{c}\left(x_{1}\right)$, $c_{1}=E_{2}^{c}\left(x_{2}\right)$; and domain-specific style representations  $s_{1}=E_{1}^{2}\left(x_{1}\right)$, $s_{2}=E_{s}^{2}\left(x_{2}\right)$. Then swap the style codes and use ${G}_2$  to produce the translated output image $x_{1 \rightarrow 2}=G_{2}\left(c_{1}, s_{2}\right)$.

\textbf{\subsection{Network architecture}}
\label{sec:network architecture}
Figure \ref{fig:network architecture} shows the network architecture of our model. It consists of a content encoder, a style encoder and a decoder.

\paragraph{Content encoder.} The content encoder consists of several convolutional layers to down-sample the input images to get high-dimension features and several basic blocks for further processing. There are many choices for basic block such as residual block \citep{He2016Deep}, residual dense block \citep{zhang2018residual}, residual in residual dense block \citep{wang2018esrgan}. Here we use the traditional residual block for simplicity and replace Batch Normalization (BN) \citep{ioffe2015batch} layer with Instance Normalization (IN) \citep{ulyanov2016instance}. For diversity, we add noise in the last two basic blocks as in \citet{DRIT}.

\paragraph{Style encoder.} The style encoder includes several strided convolutional layers, followed by an adaptive average pooling layer and a convolutional layer. We do not use IN layers in the style encoder, as IN removes the original feature mean and variance which represent important style information.

\paragraph{Decoder.} The decoder generates images from their content codes and style codes. For multi-domain translation, we also add domain class as input. Specifically, the domain class and style codes are concatenated by channel and then fed into a multi-layer perceptron (MLP). The the content codes and outputs generated by the MLP are further processed via several concatenation blocks. We equip the residual blocks with Adaptive Instance Normalization (AdaIN) \citep{huang2017arbitrary} layers whose parameters are dynamically generated by the MLP from the style codes \citep{huang2017arbitrary,ghiasi2017exploring}.
\begin{equation}
\operatorname{AdaIN}(z, \gamma, \beta)=\gamma\left(\frac{z-\mu(z)}{\sigma(z)}\right)+\beta
\end{equation}

\begin{figure}
  \includegraphics[width=0.5\textwidth]{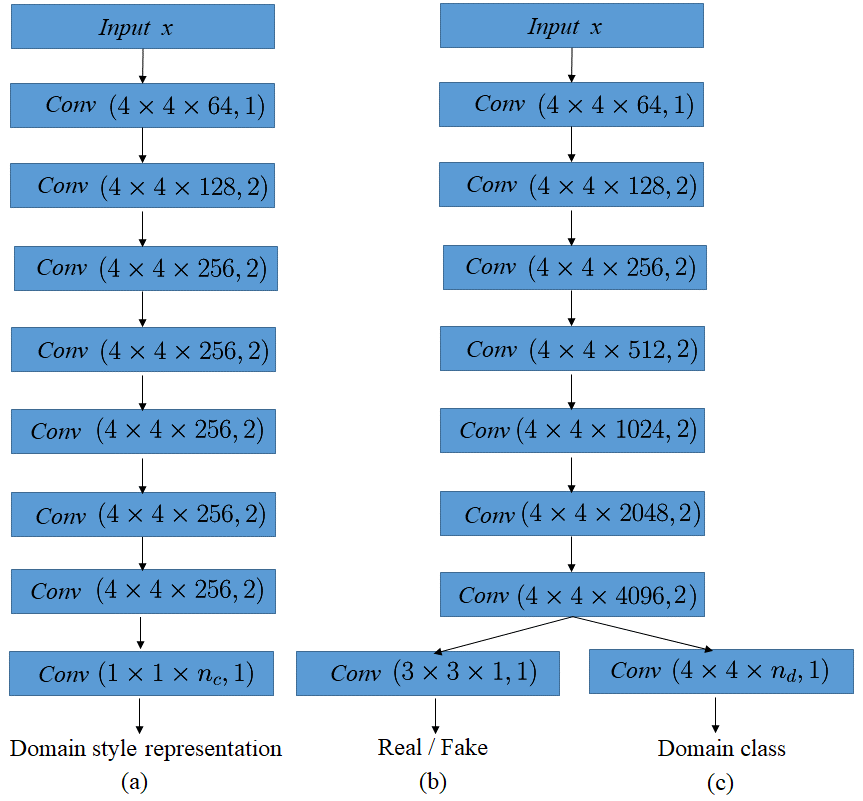}
\caption{Discriminator and domain classifier.}
\label{fig:dis}
\end{figure}

\paragraph{Discriminator and domain classifier.} The architecture of discriminator is similar with \citet{StarGAN2018}. The domain classifier is built on top of the discriminator, as shown in Figure \ref{fig:dis}. It consists of six convolution layers with kernel size $4 \times 4$ and stride $2$, following two separated convolutional branches that are implemented for discriminative output and domain class.

\paragraph{Domain style representation extractor.} The domain style representation extractor shares the same architecture with style encoder. Specifically,  it consists of one convolution layer with kernel size $4 \times 4$ and stride $1$; six convolution layers with  kernel size $4 \times 4$, stride $2$ and ReLU followed by an adaptive average pooling layer and a convolutional layer with kernel size $1 \times 1$, stride $1$.

\textbf{\subsection{Loss function}}

The loss functions are designed for unsupervised multi-domain multi-modal  image-to-image translation. For unsupervised training, we adapt image reconstruction loss and latent reconstruction loss based on cycle consistent loss. We also add constraints to improve the representations of content and style codes by self-reconstruction loss. For multi-modality, we introduce a distribution matching loss to make the style codes extracted by content encoder close to a prior Gaussian distribution. By doing this, we are able to sample style codes from prior Gaussian distribution at testing phase. Since the sampled style codes are random and stochastic, the decoder can produce diverse samples sharing the same content.
For simultaneous training of multiple different domains, we adapt a domain classification loss.

 \paragraph{Self reconstruction loss.}
  Given an image from a certain domain, we should be able to reconstruct itself after encoding and decoding. Thus the self reconstruction loss $\mathcal{L}_{\mathrm{sr}}$ can be written as
  \begin{equation}
\mathcal{L}_{\mathrm{sr}}^{x_1}=\mathbb{E}_{x_{1} \sim p\left(x_{1}\right)}\left[\left\|G_{1}\left(E_{1}^{c}\left(x_{1}\right), E_{1}^{s}\left(x_{1}\right)\right)-x_{1}\right\|_{1}\right].
\end{equation}

 \paragraph{Image reconstruction loss.}
 Given an image sampled from the data distribution, we should be able to reconstruct it after encoding and decoding. The image reconstruction loss $\mathcal{L}_{\mathrm{cc}}$ is adapted in two stages. In the pre-training of domain representations, we use image reconstruction loss to obtain a domain-specific style representation extractor $E_{\mathcal{D}_i }^{s}$ during the process of image reconstruction.
  \begin{equation}
\mathcal{L}_{\mathrm{cc}}^{x}=\mathbb{E}_{x \sim p\left(x\right)}\left[\left\|G_{1}\left(E_{1}^{c}\left(x\right), E_{\mathcal{D}_i }^{s}\left(x^{\prime}\right)\right)-x\right\|_{1}\right],
\end{equation}
where $x$ and $x^{\prime}$ are from the same domain.

%In inter-domain translation, image reconstruction loss are used for two cases which depend on whether the style is from a single image or the pre-training of domain style representation.
%In inter-domain translation, image reconstruction losses are used for two cases which depend on whether the style is from a single image or the pre-training of domain style representation.
In inter-domain translation, image reconstruction loss $\mathcal{L}_{\mathrm{cc}}$ is used for the style from a single image. The image reconstruction loss can be represented as:
 \begin{equation}
 \begin{aligned}
\mathcal{L}_{\mathrm{cc}}^{x}&=&\mathbb{E}_{x, y}\left[\left\|G_1\left(E_2^c(y'), E_1^{s}(x')\right)-x\right\|_{1}\right],\\
\mathcal{L}_{\mathrm{cc}}^{y}&=&\mathbb{E}_{x, y}\left[ \left\|G_2\left(E_1^{c}(x'), E_2^{s}(y')\right)-y\right\|_{1}\right],
\end{aligned}
\end{equation}
%where $u=G_{\mathcal{X}}\left(E_{\mathcal{Y}}^{c}(y)\right), E_{\mathcal{X}}^{s}(x) )$ and $v=G_{\mathcal{Y}}\left(E_{\mathcal{X}}^{c}(x)\right), E_{\mathcal{Y}}^{s}(y) )$.
where
 \begin{equation}
  \begin{aligned}
 x^{\prime}=&G_1\left(E_2^c(y)\right), E_1^s(x)),\\
y^{\prime}=&G_2\left(E_1^c(x)\right), E_2^s(y)).
\end{aligned}
\end{equation}

\paragraph{Disentanglement constrained loss.}
To utilize domain information and explicitly constrain the disentanglement, we propose the disentanglement loss. For style from domain style representation, the disentanglement constrained loss $\mathcal{L}_{\mathrm{dc}}$ can be represented as
 \begin{equation}
\mathcal{L}_{\mathrm{dc}}^{x}=\mathbb{E}_{x, y}\left[\left\|y^{\prime}-y^{\prime\prime}\right\|_{1}\right],\\
\end{equation}
where $y^{\prime\prime}=G_2\left(E_2^c(x), \mathcal{S}_{\mathcal{Y}} \right)$, $\mathcal{S}_{\mathcal{Y}}$ is extracted domain style.
\paragraph{Latent reconstruction loss.}
Given a latent code (style and content) sampled from the latent distribution at translation time, we should be able to reconstruct it after decoding and encoding.
\begin{equation}
\begin{aligned}
\mathcal{L}_{\mathrm{lr}}^{c_1}&=&\mathbb{E}_{c_1 \sim p\left(c_1\right), s_2 \sim q\left(s_2\right)}\left[\left\|E_2^c\left(G_2\left(c_1, s_2\right)\right)-c_1\right\|_1\right],
\\
\mathcal{L}_{\mathrm{lr}}^{s_2} &=&\mathbb{E}_{c_1 \sim p\left(c_1\right), s_2 \sim q\left(s_2\right)}\left[\left\|E_2^s\left(G_2\left(c_1, s_2\right)\right)-s_2\right\|_1\right].
\end{aligned}
\end{equation}

\paragraph{Distribution matching loss.} We adapt a distribution matching loss to make the style codes close to a prior Gaussian distribution. At testing phase, we are able to sample stochastically from prior Gaussian distribution and regard it as style code. As demonstrated in Section \ref{sec:domain adaption},  the measure of distance between two distributions can be covariance, MMD or KL divergence. Instead of implementing KL divergence as in \citet{huang2018munit} and \citet{DRIT}, here we choose the Maximum-Mean Discrepancy (MMD). We will illustrate the reasons in Section \ref{sec:ablation study}.

The distribution matching loss $\mathcal{L}_{\mathrm{dm}}$ described by MMD can be written as
\begin{equation}
\mathcal{L}_{\mathrm{dm}}=\mathbb{E}\left[D_{\mathrm{MMD}}\left(z | N(0,1)\right)\right],
\end{equation}
where
\begin{equation}
\begin{aligned}
D_{\mathrm{MMD}}(q | p) &=\mathbb{E}_{p(z), p\left(z^{\prime}\right)}\left[k\left(z, z^{\prime}\right)\right]-2 \mathbb{E}_{q(z), p\left(z^{\prime}\right)}\left[k\left(z, z^{\prime}\right)\right] \\ &+\mathbb{E}_{q(z), q\left(z^{\prime}\right)}\left[k\left(z, z^{\prime}\right)\right]
\end{aligned}
\end{equation}
 $k(\cdot, \cdot)$ can be any positive definite kernel, such as Gaussian $k\left(z, z^{\prime}\right)=e^{-\frac{\left\|z- z^{\prime}\right\|^{2}}{2 \sigma^{2}}}$.
%https://ermongroup.github.io/blog/a-tutorial-on-mmd-variational-autoencoders/#gretton2007kernel

\paragraph{Domain classification loss.}
To achieve simultaneous training of multiple domains with a single model, we assign a unique class label for each domain as in \citet{StarGAN2018}. While translating given input images $x_1$ with domain class $c_1$ to $x_2$ with $c_2$, the auxiliary domain classifier tries to distinguish images from different domains. The corresponding domain classification loss can be defined as

\begin{equation}
\begin{aligned}
\mathcal{L}_{\mathrm{cls}}^{real}=&\mathbb{E}_{x, c^{\prime}}\left[-\log D_{\mathrm{cls}}\left(c^{\prime} | x\right)\right], \\
\mathcal{L}_{\mathrm{cls}}^{fake}=&\mathbb{E}_{x, c}\left[-\log D_{\mathrm{cls}}(c | G(x, c))\right],
\end{aligned}
\end{equation}
where $D_{\mathrm{cls}}\left(c^{\prime} | x\right)$ represents a probability distribution over domain labels calculated by $D$. The goal of this term is that $D$ can correctly classify a real image $x$ to its original domain $c^{\prime}$ and $G$ tries to generate images that can be recognized as target domain $c$ by $D$.

This auxiliary domain classifier is build on top of discriminator $D$. In training phase, the domain classification loss of real images is used to optimize parameters of discriminator $D$ and the domain classification loss of fake images is used to optimize $G$.

In our experiment, the domain classifier $D_{cls}$ is pre-trained together with the domain style representation extractor $E_{\mathcal{D}_{i}}^{s}$.

\paragraph{Adversarial loss.}
For high image quality, stable training and considerable diversity as discussed in Section \ref{sec:gan}, we use the least-squares GAN proposed by \citet{mao2017least}. Thus $\mathcal{L}_{\mathrm{adv}}$ can be formulated as:

\begin{equation}
\begin{aligned}
\min _{D_{1}} \mathcal{L}_{\mathrm{adv}}\left(D_{1}\right)=&\frac{1}{2} \mathbb{E}_{\boldsymbol{x} \sim p{\mathrm(\boldsymbol{x})}}\left[\left(D_{1}(\boldsymbol{x})-b\right)^{2}\right] +\\
                                                                & \frac{1}{2} \mathbb{E}_{\boldsymbol{z} \sim p_{\boldsymbol{z}}(\boldsymbol{z})}\left[\left(D_{1}\left(G_{1}(\boldsymbol{z})\right)-a\right)^{2}\right] \\
\min _{G_{1}} \mathcal{L}_{\mathrm{adv}}\left(G_{1}\right)=  &\frac{1}{2} \mathbb{E}_{\boldsymbol{z} \sim p_{z}(\boldsymbol{z})}\left[\left(D_{1}\left(G_{1}(\boldsymbol{z})\right)-c\right)^{2}\right].
\end{aligned}
\end{equation}

\paragraph{Overall training loss.} The total training loss functions of the encoder $E$, decoder $G$ and discriminator $D$ are defined as follows:
\begin{eqnarray}
\mathcal{L}_{\mathrm{E\circ G}}^{\mathrm{total}}&=&\mathcal{L}_{\mathrm{adv}}+\lambda_{\mathrm{sr}} \mathcal{L}_{\mathrm{sr}} + \lambda_{\mathrm{cc}} \mathcal{L}_{\mathrm{cc}} +\lambda_{\mathrm{dc}} \mathcal{L}_{\mathrm{dc}}\label{eq:total_g} \nonumber\\
                                                                                &+&\lambda_{\mathrm{dm}} \mathcal{L}_{\mathrm{dm}} +\lambda_{\mathrm{lr}} \mathcal{L}_{\mathrm{lr}}+ \lambda_{\mathrm{cls}} \mathcal{L}_{\mathrm{cls}}^{fake},\\
\mathcal{L}_{D}^{\mathrm{total}}                    &=&\mathcal{L}_{\mathrm{adv}}+\lambda_{\mathrm{cls}} \mathcal{L}_{\mathrm{cls}}^{real}, \label{eq:total_d}
\end{eqnarray}
where hyper-parameters $\lambda_{sc}$, $\lambda_{cc}$, $\lambda_{dl}$, $\lambda_{dm}$, $\lambda_{lr}$ and $\lambda_{cls}$ are weights to control the importance of each term.

\paragraph{The overall process.}
The overall process is summarized in Algorithm \ref{alg: training process}. The training process consists of two phase: the domain style representation extractor training and cross-domain translation training. Both phases share almost the same network architecture and loss functions except the following differences. Since we want to learn domain style representation from each domain and adapt it to cross-domain translation as domain supervision, we select images from the same domain and swap their style codes. Ideally, the style-exchanged images should be consisted with the original ones. Only one-step translation is required to get the domain style representation. So the loss function of the domain style representation extractor training can be defined as:
\begin{eqnarray}
\mathcal{L}_{\mathrm{E\circ G}}^{\mathrm{total}}&=&\mathcal{L}_{\mathrm{adv}}+\lambda_{\mathrm{sr}} \mathcal{L}_{\mathrm{sr}}+\lambda_{\mathrm{dm}} \mathcal{L}_{\mathrm{dm}} + \lambda_{\mathrm{cls}} L_{\mathrm{cls}}^{fake},\label{eq:total_g_ds} \\
\mathcal{L}_{D}^{\mathrm{total}} &=&\mathcal{L}_{\mathrm{adv}}+\lambda_{\mathrm{cls}} \mathcal{L}_{\mathrm{cls}}^{real}, \label{eq:total_d_ds}
\end{eqnarray}
where hyper-parameters $\lambda_{\mathrm{sc}}$, $\lambda_{\mathrm{dm}}$, $\lambda_{\mathrm{cls}}$ are weights to control the importance of each term.

Thus we get the domain style representation extractor $E_{\mathcal{D}_{i}}^{s}$. It mainly used in image reconstruction loss to constrain feature disentanglements.

\begin{algorithm}[t]
%https://www.latex4technics.com/?note=1v2f
\let\oldnl\nl% Store \nl in \oldnl
\newcommand{\nonl}{\renewcommand{\nl}{\let\nl\oldnl}}% Remove line number for one line
\makeatother
%\DontPrintSemicolon
\textbf{Input}: $N$ different domains $\mathcal{D}_{k} \ \forall k \in[N]$, batch size $N$, learning rate $\eta$;\\
\nonl \textbf{Stage 1: domain style representation extractor training}\\
Randomly initialize the parameters $\Theta_{E}$ of domain representation extractor $E_{\mathcal{D} }^{s}$;\\
Randomly select one domain $\mathcal{D}_{k}$, $k \in[N]$.  Get a mini-batch of data $\mathcal{D}_{k}$ satisfying $D_{k} \subset \mathcal{D}_{k}$ and $\left|D_{k}\right|=K$;\\
Update the network as follows:
$\Theta_{E \circ G} \leftarrow \Theta_{E \circ G}-\eta \nabla_{\Theta} \Theta_{E \circ G} \ell_{\mathrm{E \circ G}}^{\text {total }}\left(D_{\mathcal{S}}\right)$
$\Theta_{D} \leftarrow \Theta_{D}-\eta \nabla_{\Theta_{D}} \ell_{D}^{\mathrm{total}}\left(D_{\mathcal{S}}\right)$\\
where $\ell_{\mathrm{E \circ G}}^{\text {total }}\left(D_{\mathcal{S}}\right)$ and $\ell_{D}^{\text {total }}\left(D_{\mathcal{S}}\right)$ are defined in Eqn. \ref{eq:total_g_ds} and Eqn. \ref{eq:total_d_ds}, respectively.\\
 Repeat step $3$ and step $4$ until convergence.\\
\nonl \textbf{Stage 2: cross-domain translation training}\\
Randomly initialize the parameters $\Theta_{E \circ G}$ of content encoder $E^{c}$, style encoder $E^{s}$, decoder $G$ and parameters $\theta_{G}$ of discriminator $D$;\\
Randomly select two different domains $\mathcal{D}_{A}, \mathcal{D}_{B}, A, B \in[N]$ . For each selected domain $\mathcal{D}_{l}$ where $l \in\{A, B\}$, get a minibatch of data $D_{l}$ satisfying $D_{l} \subset \mathcal{D}_{l}$ and $\left|D_{l}\right|=K$.\\
\If{Training}
{
Update the parameters as follows:
$\Theta_{E \circ G} \leftarrow \Theta_{E \circ G}-\eta \nabla_{\Theta} \Theta_{E \circ G} \ell_{\mathrm{E \circ G}}^{\text {total }}\left(D_{A}\right)$
$\Theta_{D} \leftarrow \Theta_{D}-\eta \nabla_{\Theta_{D}} \ell_{D}^{\mathrm{total}}\left(D_{A}\right)$
}
where $\ell_{\mathrm{E \circ G}}^{\text {total }}\left(D_{A}\right)$ and $\ell_{D}^{\text {total }}\left(D_{A}\right)$ are defined in Eqn. \ref{eq:total_g} and Eqn. \ref{eq:total_d}, respectively.\\
Repeat step $7$ and step $10$ until convergence.
\caption{{\sc Training process.}}
\label{alg: training process}
\end{algorithm}

\section{Experiment}
\label{sec:experiment}

\begin{figure*}
  \includegraphics[width=1.0\textwidth]{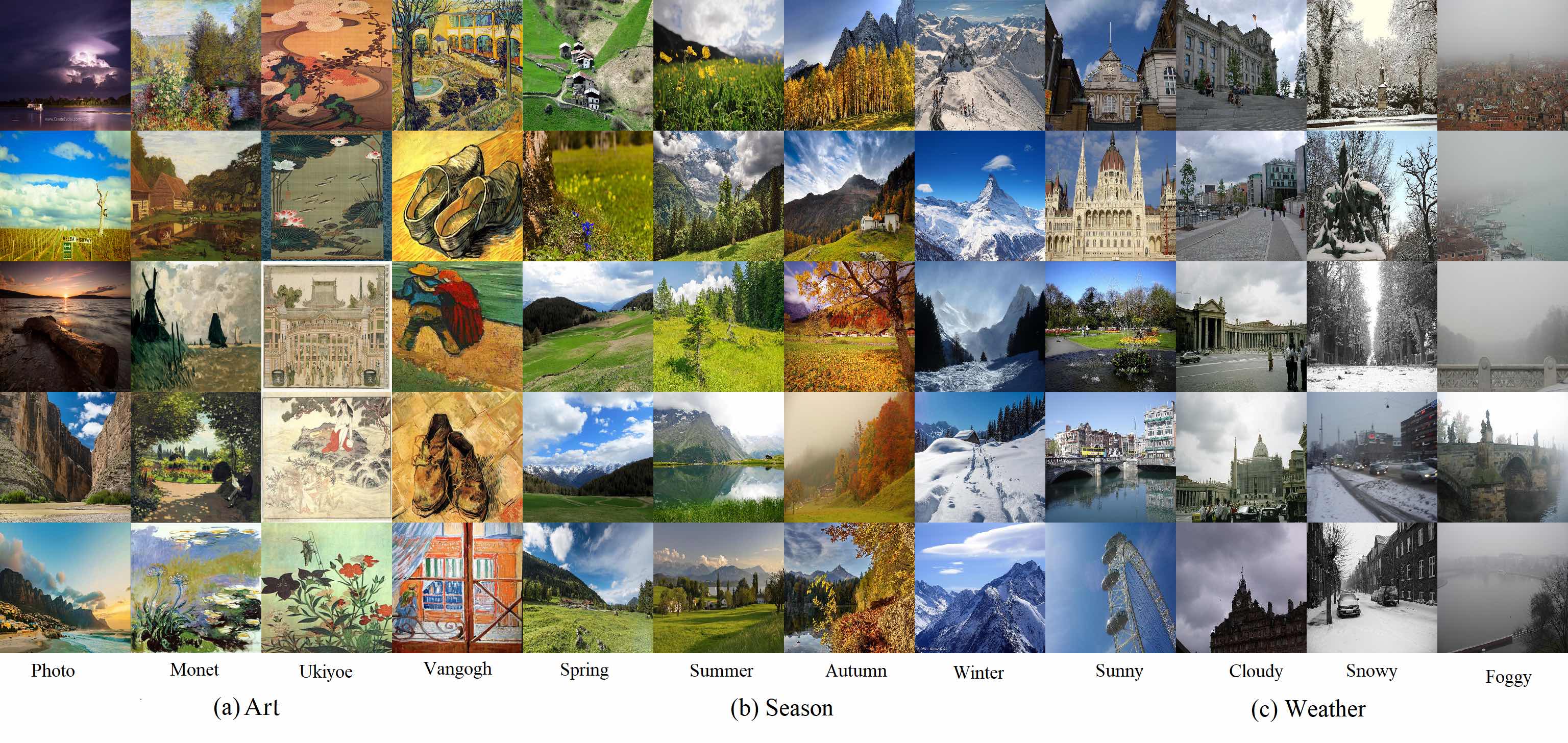}
\caption{Samples from datasets. We mainly use three multi-domain datasets for experiments: Art, Season and Weather. Each contains four domains.}
\label{fig:data_samples}
\end{figure*}

\textbf{\subsection{Experiment Settings}}
For training, we adapt Adam optimizer with a batch size $8$, a learning rate of $0.0001$ with exponential decay rates $\beta_{1}=0.5$, $\beta_{2}=0.999$. We resize all input images into $216 \times 216$ in experiments. The hyper-parameters are set as $\lambda_{\mathrm{sr}}=10$, $\lambda_{\mathrm{cc}}=10$, $\lambda_{\mathrm{dm}}=0.01$, $\lambda_{\mathrm{lr}}=10$. And $\lambda_{\mathrm{cls}}$ of $G$ is $5.0$, $\lambda_{\mathrm{cls}}$ of $D$ is $1.0$. We don't implement domain supervision if training data are paired.

\textbf{\subsection{Datasets}}

We use three multi-domain datasets for experiments: art, weather, season. Notice that all images in these datasets are not paired.

\textbf{Art}: This dataset contains four domains: real images, Monet, Ukiyo-e and Van Gogh. These art images can be download from Wikiart \footnote{https://www.wikiart.org/} and the real photos are from Flickr with tags \textit{landscape} and \textit{landscapephotography}. We use the monet2photo, vangogh2photo, ukiyoe2photo and cezanne2photo datasets collected by \cite{CycleGAN2017}.

\textbf{Weather}: This dataset contains four domains: sunny, cloudy, snowy, and foggy, which is randomly selected from the Image2Weather \citep{chu2017camera}.

\textbf{Season}: This dataset consists of approximately $6,000$ images of the Alps mountain range scraped from Flickr. The original dataset collected by \citet{anoosheh2018combogan} categorizes photos individually into four seasons based on the provided timestamp of when it was taken. But this lead to many misclassifications. We revise each category by deleting ambiguous images or removing misclassified images to the right category to make them more distinguishable.

Since \citet{zhu2017toward} need paired data for training, we evaluate multi-modality on \textbf{edges $\rightarrow$ shoes} and \textbf{edges $\rightarrow$ handbags}. The edges $\rightarrow$ shoes dataset contains 50k training images from UT Zappos50K dataset \citep{DBLP:conf/cvpr/YuG14}. The edges $\rightarrow$ handbags dataset contains 137K Amazon Handbag images from \citet{zhu2016generative}. Edges are computed by HED edge detector \citep{xie15hed} and post-processing. Both datasets can be downloaded at CycleGAN \citep{CycleGAN2017} website\footnote{https://github.com/junyanz/pytorch-CycleGAN-and-pix2pix}.

Samples from these three datasets are visually demonstrated in Figure \ref{fig:data_samples} to describe their styles. And Table \ref{tab:datasets} describes domain information and corresponding number of training data.
\begin{table}[t]
  \centering
 \caption{Features of each datasets.}
\label{tab:datasets}
{
\begin{tabular}{cccccc}
\toprule
Art                         & Num.   	  & Weather 	& Num.           & Season        & Num. 	 \\
\midrule
Photos                   &	2853 	   & Sunny	    &70601 	        & Spring	        & 1382	\\
Monet                    &	1074	   & Cloudy     & 45662          & Summer	    &1512		\\
Van Gogh              & 	401	       & Foggy      & 357		        & Autumn	    &1606	    \\
Ukiyo-e                 & 1433	        & Snowy    &1252	            & Winter	        &993	    \\
\bottomrule
\end{tabular}
}
\end{table}

\textbf{\subsection{Baselines}}
We perform the evaluation on the following baseline approaches:

\textbf{BicycleGAN.}
BicycleGAN \citep{zhu2017toward} is the first image-to-image translation model that aims to generate continuous and multi-modal output images. However, it needs paired images for training.

\textbf{DRIT} \citep{DRIT} and \textbf{MUNIT} \citep{huang2018munit}
propose to simultaneously generate diverse outputs given the same
input image without requirement of  pair supervision via disentangled representations. It's designed for translation between two domains.

\textbf{StarGAN.}
StarGAN \citep{StarGAN2018} aim to handle scalability of unsurprised image-to-image translation problems. It uses one generator and discriminator in common for all domains by adding domain labels. The generator requires images and the desired domain label specifying the target domain as inputs, and the discriminator is trained to classify the domain labels of generated images and judge whether it's real or fake. By doing this, it's able to take any number of domains as input. However, the model was just applied on task of face attribute translation in original paper. It didn't validate on datasets with various categories. Furthermore, it didn't pay attention on the problem of multi-modality.

\textbf{DosGAN.}
 DosGAN \citep{lin2019unpaired} share the similar idea of \citep{StarGAN2018} to achieve simultaneous training for multi-domains. It farther introduce the domain supervision, which uses domain-level information as supervision and pre-trains a classifier to predict which domain an image is from. The authors believe that the classifier should carry rich domain signal. Therefore, the output of the second-to-last layer of this classifier can be leveraged to extract the domain features of an image. Still, it follows the same drawback to diversity with \citet{StarGAN2018}.

\textbf{ComboGAN.}
Different from \citet{StarGAN2018} and \citet{lin2019unpaired}, \citet{anoosheh2018combogan} don't use domain labels to achieve simultaneous training for multi-domains. Instead, it uses $n$ generators and discriminators for translations among $n$ domains. Specifically, it divides each generator network in half, labeling each one as encoders and decoders, respectively, and then assigns an encoder and decoder to each domain.

Since that those methods are designed for different purposes, we conduct comparisons on two criterions. For simultaneous training, we compare our approach with \citet{StarGAN2018}, \citet{lin2019unpaired} and \citet{anoosheh2018combogan}. For multi-modality, we compare our method with \citet{zhu2017toward}, \citet{DRIT} and \citet{huang2018munit}.

\begin{figure*}[t]
  \includegraphics[width=1.0\textwidth]{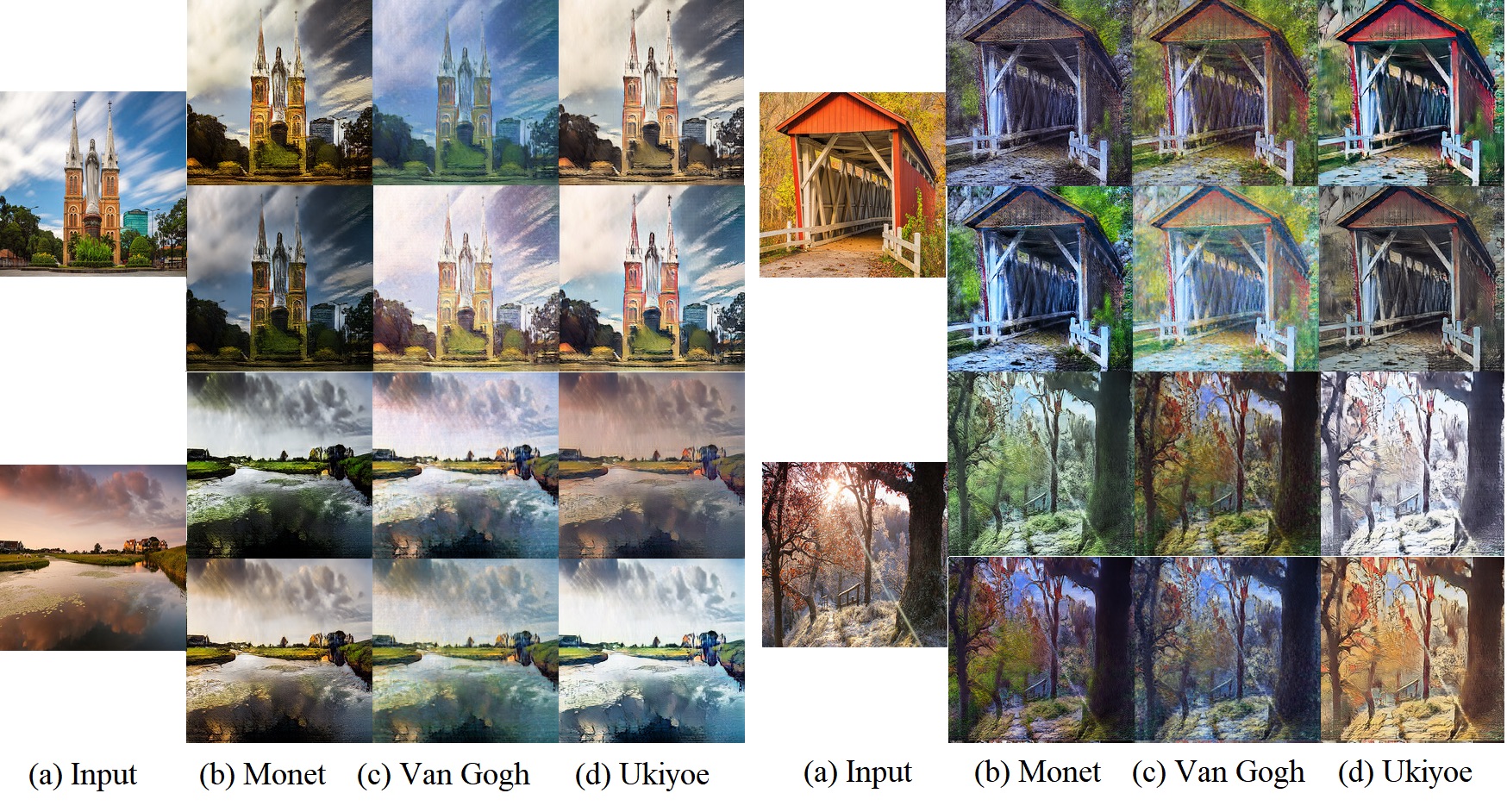}
\caption{Multi-domain multi-modal image translation results on Art. The Art dataset contains four domain: real image, Monet, Van Gogh and Ukiyoe. Better look by zooming in.}
\label{fig:art_result}
\end{figure*}

\begin{figure*}[t]
  \includegraphics[width=1.0\textwidth]{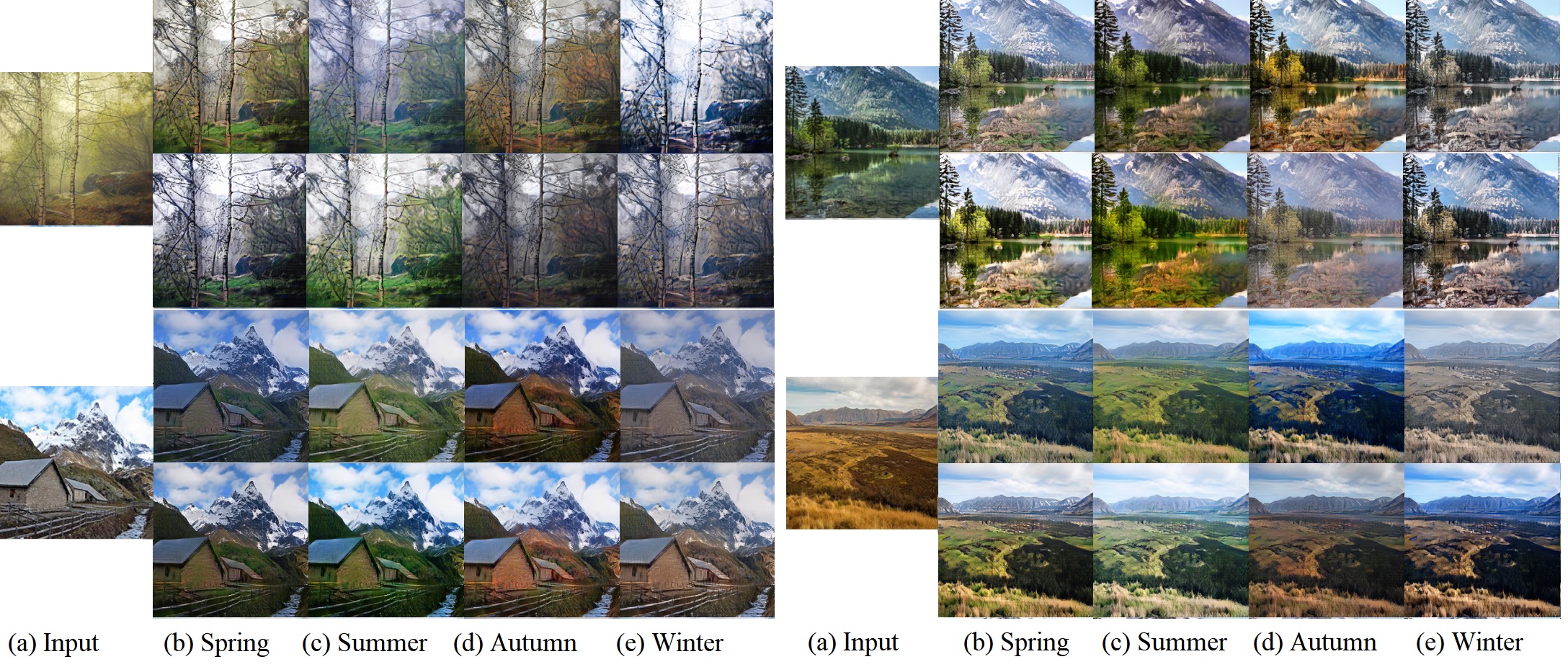}
\caption{Multi-domain multi-modal translation result on Season. The Season dataset contains four domain: spring, summer, autumn and winter. Notice that all these image are generated via one training process. Better look by zooming in.}
\label{fig:season_result}
\end{figure*}

\textbf{\subsection{Evaluation Metrics}}
We use the Fr$\acute{e}$chet Inception distance and LPIPS Distance to evaluate the quality and diversity of the generated images.

%We use the Inception Score, Fr$\acute{e}$chet Inception distance and LPIPS Distance to evaluate the quality and diversity of the generated images.

\textbf{LPIPS Distance.} Similar to \citet{zhu2017toward}, we use Learned Perceptual Image Patch Similarity (LPIPS) metric \citep{zhang2018unreasonable} to measure translation diversity. LPIPS Distance is calculated by a weighted $\mathcal{L}_{2}$ distance between deep features of randomly-sampled translation results from the same input. It has been shown to correlate well with human perceptual similarity.

%\textbf{Inception Score.} Many image generation methods adapt the Inception Score (IS) \citep{salimans2016improved} as metric to evaluate the quality of the generated samples. Though it has been demonstrated to be flawed in some recent works \citep{barratt2018note}, we report the IS to enable comparison with existing methods.

\textbf{FID score.} Fr$\acute{e}$chet Inception distance (FID) \citep{heusel2017gans} is a measure of similarity between two datasets of images. It was shown to correlate well with human judgement of visual quality and is most often used to evaluate the quality of samples of Generative Adversarial Networks. FID is calculated by computing the Fr$\acute{e}$chet Inception distance between two Gaussians fitted to feature representations of the Inception network.

%\textbf{Kernel-Inception Distance.} Kernel-Inception Distance (KID) \citep{binkowski2018demystifying} measures the dissimilarity between two probability distributions $P_r$ and $P_g$ using samples drawn independently from each distribution.

%\begin{table*}[t]
%  \centering
% \caption{Evalution metircs.}
%\label{tab:evalution metircs}
%\begin{tabular}{lp{12cm}p{3cm}p{2cm}}%{ccc}
%\toprule
%Name	                                           &Description	                                                                                                                                                                       &Performance score \\
%\midrule
%LPIPS  &Distance in feature space with linear weights to better match human perceptual judgments.                                                     &Higher is better. \\
%IS                           &KL-Divergence between conditional and marginal label distributions over generated data.	                                       &Higher is better. \\
%FID	&Wasserstein-2 distance between multi-variate Gaussians fitted to data embedded into a feature space.	                               &Lower is better. \\
%KID	&Measures the dissimilarity between two probability distributions Pr and Pg using samples drawn independently from each distribution.	&Lower is better. \\
%\bottomrule
%\end{tabular}
%\end{table*}

\textbf{\subsection{Qualitative evaluation}}
\label{sec:qualitative evaluation}

\begin{figure*}
  \includegraphics[width=1.0\textwidth]{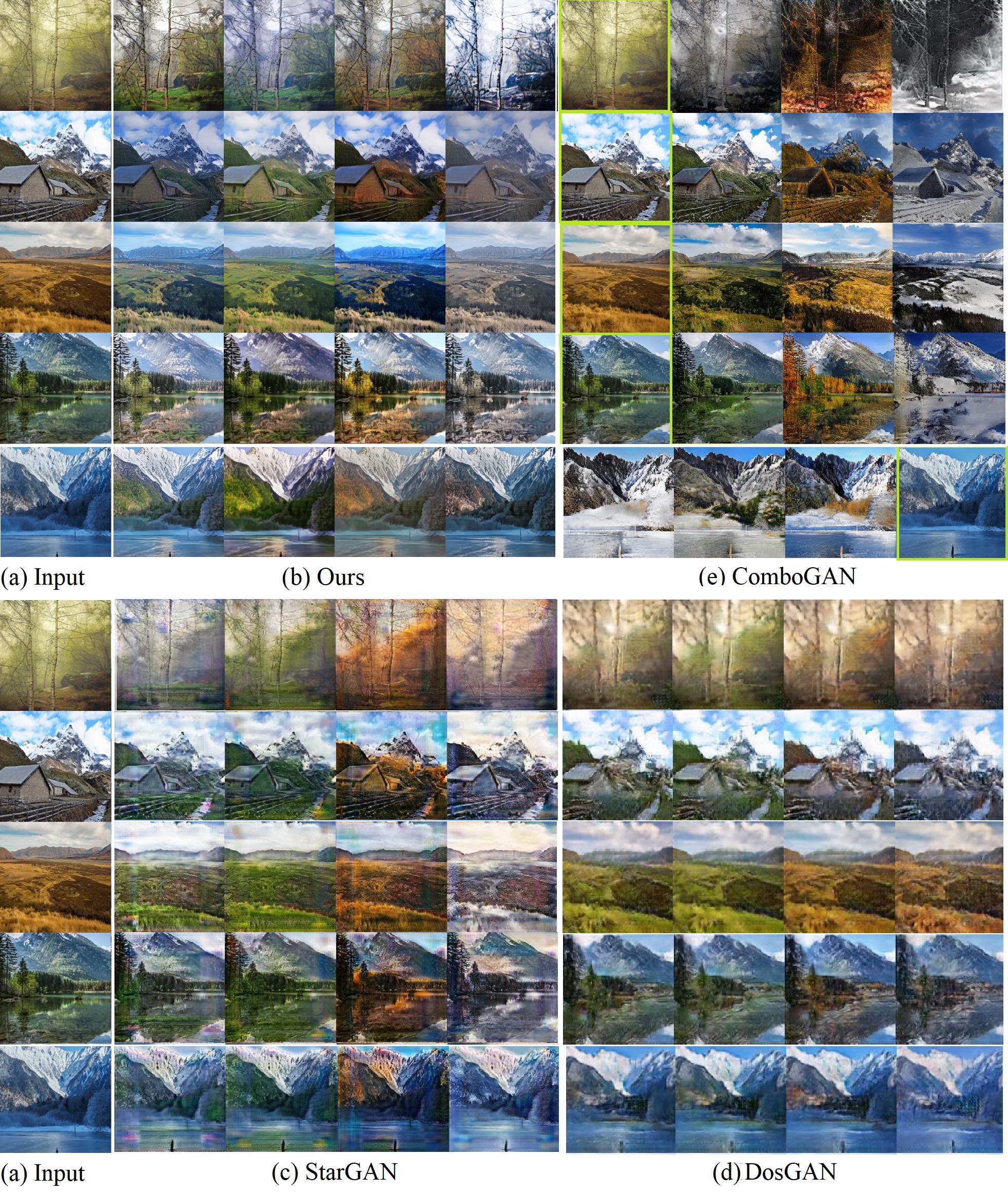}
\caption{Results of StarGAN, DosGAN, ComboGAN and ours on Season dataset. Images in the first column are input images randomly selected from the four seasons. Following are results generated by ours, StarGAN, DosGAN and ComboGAN. For each method, the four columns are arranged successively as spring, summer, autumn and winter. Better look by zooming in.}
\label{fig:compare_season_results}
\end{figure*}

We conduct comparisons on two criterions. For simultaneous training, we compare our approach with \citet{StarGAN2018}, \citet{lin2019unpaired} and \citet{anoosheh2018combogan} on season dataset. For multi-modality, we compare our method with \citet{zhu2017toward}, \citet{DRIT} and \citet{huang2018munit} on edges $\rightarrow$ shoes and edges $\rightarrow$ handbags datasets. Qualitative comparison of simultaneous multi-domain translation with baselines on Season dataset are demonstrated in Figure \ref{fig:compare_season_results}. The results produced by \citet{StarGAN2018} all have obvious artifacts. \citet{lin2019unpaired} generate fewer artifacts than \citet{StarGAN2018}. However, the results are still unpleasing and lack diversity for different seasons. In most cases, the translated spring and summer images are almost indistinguishable. All four translated season images are even nearly the same in the last row of \citet{lin2019unpaired}. \citet{anoosheh2018combogan} generate better results in both realism and diversity than the aforementioned two methods. However, it needs $8$ generators and$4$ discriminators to achieve conversion of four seasons between any two.

\begin{figure}
  \includegraphics[width=0.5\textwidth]{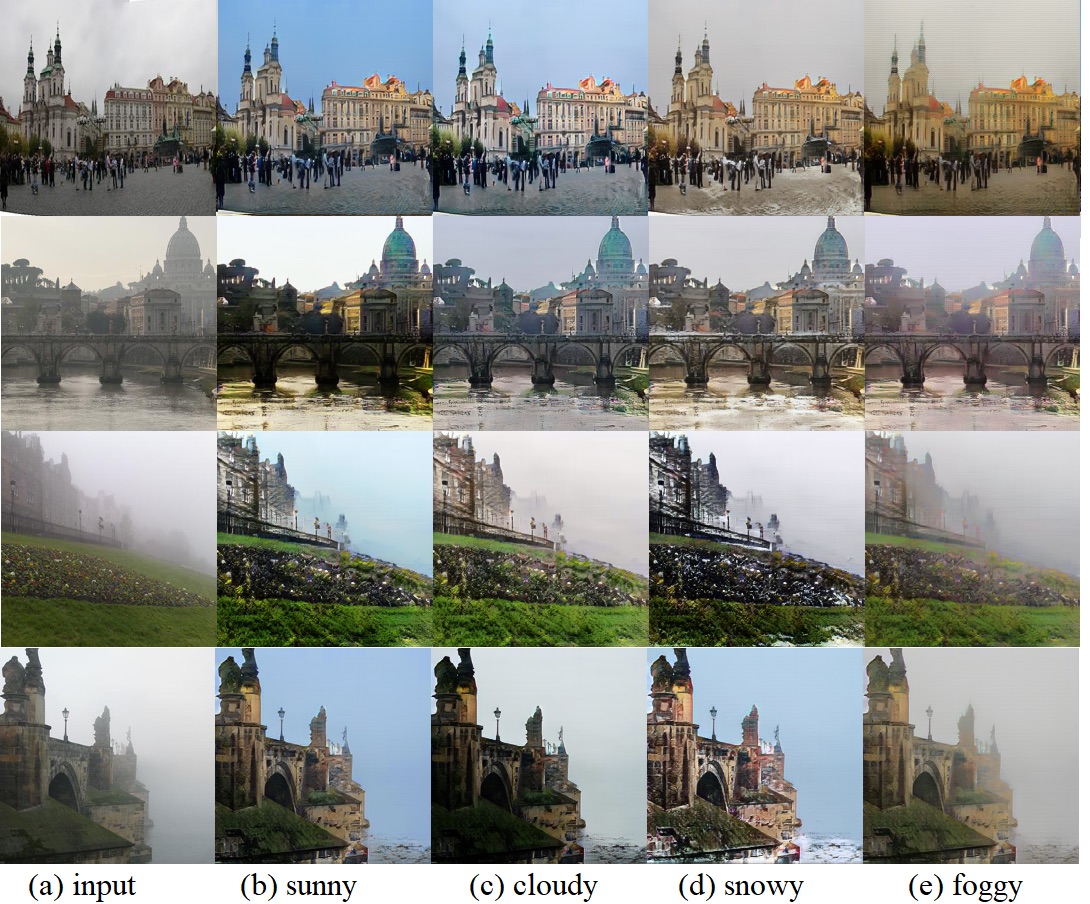}
\caption{Weather result. The images in the first row demonstrate that our method can handle images with
complex and elaborate structures. The rest images show its potential capacities to the image defogging tasks. Better look by zooming in.}
\label{fig:weather_result}
\end{figure}

Compared with baseline methods, our approach generates high-quality images which are more photo-realistic and diverse. The green blocks in Figure \ref{fig:compare_season_results} represent real input images. Those real images can still be easily told apart from the generated ones translated by \citet{anoosheh2018combogan}. In terms of realism,the real input images are indistinguishable from the four images generated by our method while in terms of diversity, the four season images can be easily classified into corresponding category. More results of our method on art, season and weather translation can be found in Figure \ref{fig:art_result}, Figure \ref{fig:season_result} and Figure \ref{fig:weather_result}.

Figure \ref{fig:weather_result} shows the results of our methods conducted on weather dataset. The images in the first row demonstrate that our method can handle images with complex and elaborate structures. The rest images show its potential capacities to the image defogging tasks.

\begin{figure}
  \includegraphics[width=0.5\textwidth]{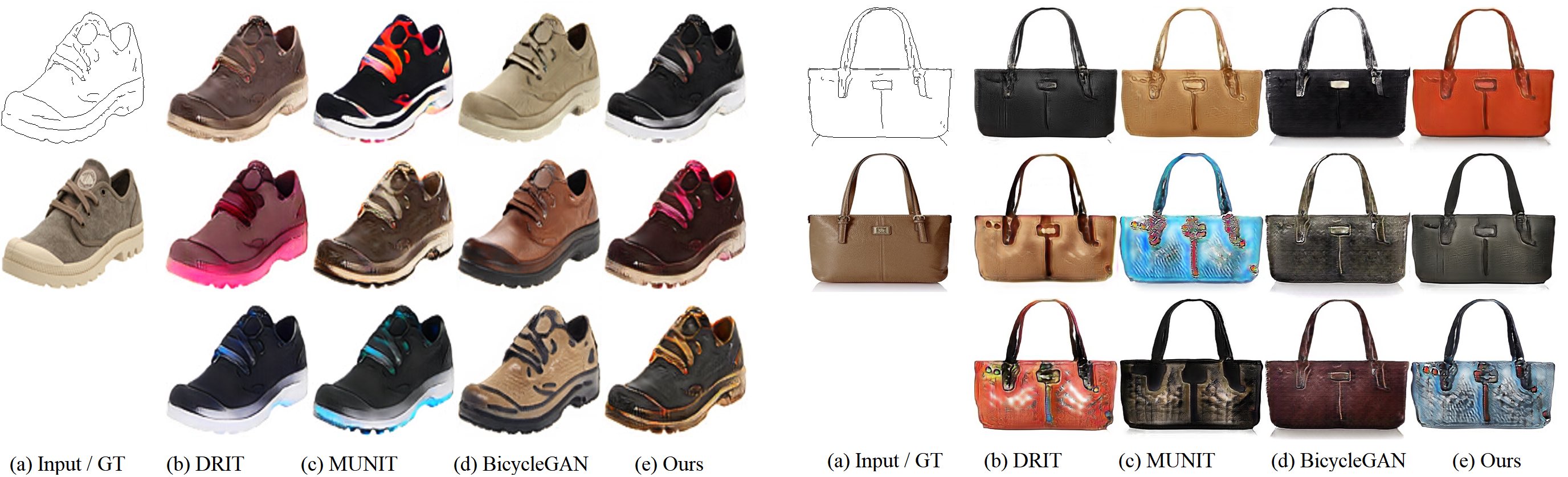}
\caption{Qualitative comparison on edges $\rightarrow$ shoes and edges $\rightarrow$ handbags. The first column shows the input and ground truth image. Each following column shows three random possible outputs from corresponding method. Better look by zooming in.}
\label{fig:shoes_bags_stitch_result}
\end{figure}

Figure \ref{fig:shoes_bags_stitch_result}  shows the results of qualitative comparison on edges $\rightarrow$ shoes and edges $\rightarrow$ handbags.

\textbf{\subsection{Quantitative evaluation}}
\label{sec:quantitative evaluation}

We conduct the quantitative evaluation on the realism and diversity of season cross-domain translation \citep{anoosheh2018combogan}.

\begin{figure}
  \includegraphics[width=0.5\textwidth]{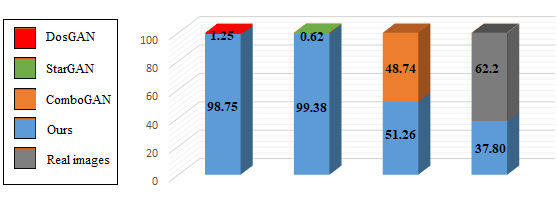}
\caption{Realism preference results. We conduct a user study to ask people to select a more realistic one between ours and \citet{StarGAN2018}, \citet{lin2019unpaired}, \citet{anoosheh2018combogan}. The number indicates the percentage of preference on the pairwise comparisons. We use the season translation for this experiment.}
\label{fig:subjective_result}
\end{figure}

For realism, we conduct a user study using pairwise comparisons. Given a pair of images sampled from real images and translation outputs generated from different methods, users need to answer two questions: ``Which image of this pair is more realistic?'' and ``Which season is this image?'' They are given unlimited time to select their preferences. For each comparison, we randomly generate $100$ questions and each question is answered by $30$ different persons. Table \ref{tab:quantitative evaluation_subjective} show the results of fooling rate and season classification accuracy. \citet{anoosheh2018combogan} get the highest fooling rate of $47.33\%$ and ours rank the second highest. Notice that \citet{anoosheh2018combogan} use several encoders and decoders to achieve this and our method only use one model. For season classification accuracy, since many images in Season dataset are too ambiguity to classify it into a certain season, the classification accuracy of the real images is like random guess, $48.9\%$. But the image-to-image translation methods tend to learn the general properties, the generated images are endowed with more distinguishable properties of certain season. \citet{lin2019unpaired} and \citet{anoosheh2018combogan} get higher classification accuracy that real images, i.e., $54.2\% $ and $55.6\%$. And ours achieve the highest accuracy of $65.8\%$, which means the domain-specific styles are better captured by our proposed method. Figure \ref{fig:subjective_result} demonstrate the realism preference results. We conduct another user study to ask people to select a more realistic one between ours and \citet{StarGAN2018}, \citet{lin2019unpaired}, \citet{anoosheh2018combogan}, real images. The number indicates the percentage of preference on the pairwise comparisons. We use the season translation for this experiment.

For diversity, similar to \citet{zhu2017toward}, we use the LPIPS metric to measure the similarity among images. Additionally, we implement FID to acquire perceptual scores. We compute the distance between $1000$ pairs of randomly sampled images translated from $100$ real images. As shown in Table \ref{tab:quantitative evaluation}, our method achieves the lowest FID scores, which means that our method produces the best results in both high-level similarity and perceptual judgement, and the highest LPIPS scores, which means the most diverse results.

%\textbf{(Not finished. After getting edges $\rightarrow$ shoes quantitative evaluation.)}
As \citet{zhu2017toward} need paired data for training, we evaluate multi-modality on edges $\rightarrow$ shoes and edges $\rightarrow$ handbags. We use the LPIPS and FID metric to compare our method with the existing state-of-the-art method, i.e., \citet{zhu2017toward,DRIT,huang2018munit}. As shown in Table \ref{tab:quantitative evaluation of multi-modality}, our method outperforms the supervised method \citep{zhu2017toward} and produce comparable results with other unsupervised methods \citep{zhu2017toward,DRIT,huang2018munit}.

\begin{table}[t]
  \centering
 \caption{Perfomance as the LPIPS and FID on the Season dataset. The \textbf{best} and \underline{second} best results are highlighted in each column. For details refer to Section \ref{sec:quantitative evaluation}.}
\label{tab:quantitative evaluation}
{
\begin{tabular}{cccc}
\toprule
Method                                                 &LPIPS   &FID   \\%& KID\\
%\midrule
%Real photos                                            & -  	      & - & -   \\
\midrule
\citet{StarGAN2018}                            &0.4273  	      & 221.7 \\%& -    \\
\citet{lin2019unpaired}                        &0.2503	  	      & 145.3 \\%& -   \\
\citet{anoosheh2018combogan}          &\underline{0.4349}	  	      & \underline{109.99} \\%& -    \\
Ours                                                       &\textbf{0.4810}    &\textbf{73.47}  \\%& -   \\%(mean,stddev)(0.504,0.69)
\bottomrule
\end{tabular}
}
\end{table}

\begin{table}[t]
  \centering
 \caption{Perfomance as the Fooling Rate and Season Classification Accuracy on the Season dataset. We conduct the user study to select results that are more realistic through pairwise comparisons and distinguish which season of an image is. The number indicates the percentage of preference on that comparison pair. The \textbf{best} and \underline{second} best results are highlighted in each column. For details refer to Section \ref{sec:quantitative evaluation}.}
\label{tab:quantitative evaluation_subjective}
{
\begin{tabular}{cccc}
\toprule
Method                                                 &Fooling Rate                    &Accuracy  \\
\midrule
Real photos                                            & -  	                                  &48.9\%\\
\midrule
\citet{StarGAN2018}                            &5.3\%	                          &41.3\% \\
\citet{lin2019unpaired}                        &27.2\%  	                      &54.2\%\\
\citet{anoosheh2018combogan}          &\textbf{47.33\%}          &\underline{55.6\%}\\
Ours                                                       &\underline{37.8\%}      &\textbf{65.8\%}\\
\bottomrule
\end{tabular}
}
\end{table}

\begin{table}[t]
  \centering
 \caption{Diversity. We use the LPIPS and FID metric to measure the diversity of generated images on the edges $\rightarrow$ shoes and edges $\rightarrow$ handbags. The \textbf{best} and \underline{second} best results are highlighted in each column.}
\label{tab:quantitative evaluation of multi-modality}
{
\begin{tabular}{ccccc}
\toprule
    \multirow{2}{*}{Method} &
    \multicolumn{2}{c}{edges $\rightarrow$ shoes} &
    \multicolumn{2}{c}{edges $\rightarrow$ handbags}\\
                                                  &LPIPS                         &FID                           &LPIPS                          &FID  \\
%\midrule
%Real photos                                            & -  	      & - & - & -    \\
\midrule
\citet{zhu2017toward}            &0.2443	  	                &115.87                      &0.3180                            &184.56 \\
\citet{DRIT}                             &0.2631	  	                &\textbf{62.67}         & \underline{0.3760}      &\underline{90.89}    \\
\citet{huang2018munit}          &\textbf{0.2652}     	&65.87                        & \textbf{0.3820}            & 91.43   \\
Ours                                           &\underline{0.2639}	&\underline{64.46}   & 0.3759                            &\textbf{89.19} \\
\bottomrule
\end{tabular}
}
\end{table}

\textbf{\subsection{Ablation study}}
\label{sec:ablation study}

\paragraph{The effect of domain supervision.}
As illustrated in Section \ref{sec:domain supervision}, the de-blurred images of \citet{DRIT} have different face contour with original ones, which means that the attribute extractor has not only learned blur distortion pattern but also recognize some content representations such as face contour as attribute. It might be caused by the ambiguous and implicit disentanglement of content and style. Thus we introduce explicit domain-constraint for disentanglement of content and style to better utilize domain information and explicitly constrain the disentanglement learning.

\begin{figure}[ht]
  \includegraphics[width=0.5\textwidth]{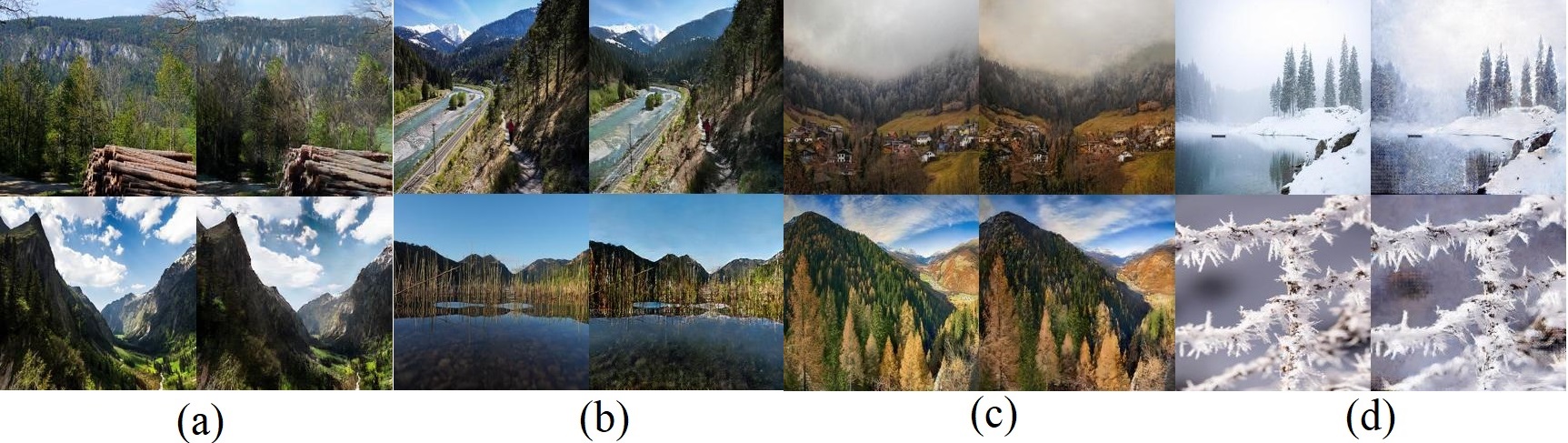}
\caption{Results of intra-domain supervision. Better look by zooming in.}
\label{fig:intra-domain}
\end{figure}

\begin{figure}[h]
  \includegraphics[width=0.5\textwidth]{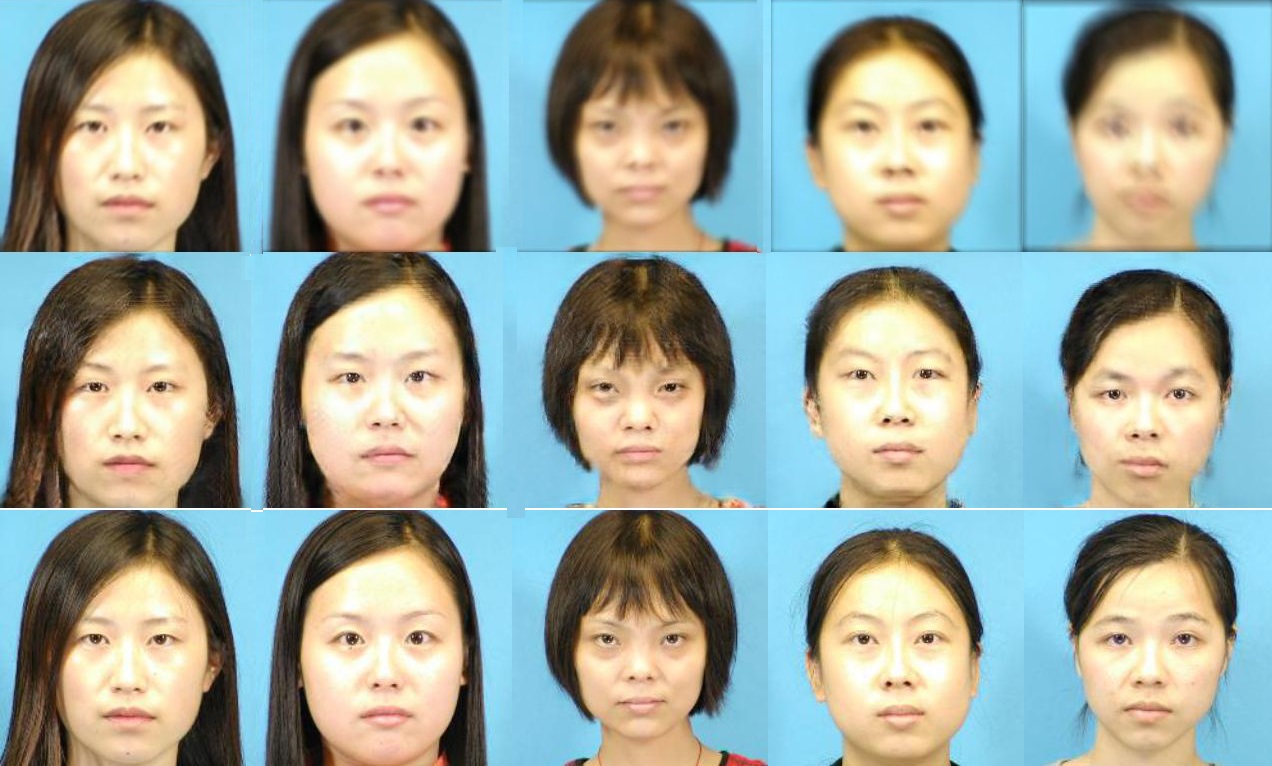}
\caption{Results of adapting our method for image de-blurring. Better look by zooming in.}
\label{fig:deblur_ours}
\end{figure}

Figure \ref{fig:intra-domain} shows the style-swapped reconstruction results of intra-domain translation. It only shows that the pre-trained model could reconstruct style-swapped images from the same domain but fail to proof the domain supervision help the explicit disentanglement learning of content and style. To further validate the effectiveness of domain supervision, we adapt our method for image de-blurring task, and compare with the results of \citet{DRIT} in Figure \ref{fig:drit_deblur}. The blurred images are generated using the same method as in \citet{yu2018crafting} and CUFS dataset \citep{DBLP:journals/pami/WangT09}. The results of image de-blurring after adding proposed disentanglement loss are shown in Figure \ref{fig:deblur_ours}. Compared with Figure \ref{fig:drit_deblur}, the generated image are consistent with the original except the blur distortion are removed.

Furthermore, we also found that perceptual loss \citep{Johnson2016Perceptual} can achieve similar disentangled constraint. The perceptual loss is based on perceptual similarity, which is often computed as the distance of two activated features in a pre-trained deep neural network between the output and the reference image:
%\citet{huang2018munit} propose a modified version of perceptual loss which is more domain-invariant. This modified perceptual loss, which they call domain-invariant perceptual loss, first performs Instance Normalization (IN) \citep{ulyanov2016instance} on the VGG features to remove the feature mean and variance. The obtained features contain domain-specific information \citep{huang2017arbitrary,li2016revisiting}, then computes the distance of the obtained features.
\begin{equation}
\mathcal{L}_{\text {percep}}=\mathbb{E}\left[\sum_{i} \frac{1}{N_{i}}\left\|\phi_{i}\left(\mathrm{I_{gt}}\right)-\phi_{i}(\mathrm{I_{pred}})\right\|_{1}\right],
\end{equation}
where $\phi_{i}$ donates the feature maps of the pre-trained VGG-19 network.

The perceptual loss and disentanglement restrained loss constrain the learning of disentangled representations in different aspects. The former, which is image-level, forces the generated images sharing the same content with the input ones. The latter, which is collection-level, restrains the style encoder from learning any content of images.

\paragraph{The measure of distributions.}
Many criteria can be used to estimate the distance between distributions. Kullback-Leibler (KL) divergence may be the most widely used in practice:
\begin{equation}
L_{\mathrm{dm}}=\mathbb{E}\left[D_{\mathrm{KL}}\left(\left(z\right) | N(0,1)\right)\right],
\end{equation}
where $D_{\mathrm{KL}}(p | q)=-\int p(z) \log \frac{p(z)}{q(z)} \mathrm{d} z$.

%However, many of these criteria are parametric. In many cases, we require an intermediate density estimate. To avoid such a non-trivial task, a non-parametric distance estimate between distributions is more desirable.
%https://ermongroup.github.io/blog/a-tutorial-on-mmd-variational-autoencoders/#chen2016variational
However, researchers have noticed that KL divergence might be too restrictive \citep{bowman2015generating,sonderby2016ladder,chen2016variational,binkowski2018demystifying}. Sometimes it failed to learn any meaningful latent representation. Several methods \citep{bowman2015generating,sonderby2016ladder,chen2016variational} try to alleviate this problem, but do not completely solve the issue.
\citet{borgwardt2006integrating} propose the Maximum Mean Discrepancy (MMD) as a relevant criterion for comparing distributions based on the Reproducing Kernel Hilbert Space (RKHS). It's a framework to quantify the distance of two c by calculating all of their moments. It can be efficiently adapted using kernel trick.
\begin{equation}
\begin{aligned}
D_{\mathrm{MMD}}(q \| p) &=\mathbb{E}_{p(z), p\left(z^{\prime}\right)}\left[k\left(z, z^{\prime}\right)\right]-2 \mathbb{E}_{q(z), p\left(z^{\prime}\right)}\left[k\left(z, z^{\prime}\right)\right] \\ &+\mathbb{E}_{q(z), q\left(z^{\prime}\right)}\left[k\left(z, z^{\prime}\right)\right],
\end{aligned}
\end{equation}
where $k(\cdot, \cdot)$ can be any positive definite kernel, such as Gaussian $k\left(z, z^{\prime}\right)=e^{-\frac{\left\|z-z^{\prime}\right\|^{2}}{2 \sigma^{2}}}$. Therefore, the distance between distributions of two samples can be well-estimated by the distance between the means of the two samples mapped into a RKHS.

\begin{figure}[h]
  \includegraphics[width=0.5\textwidth]{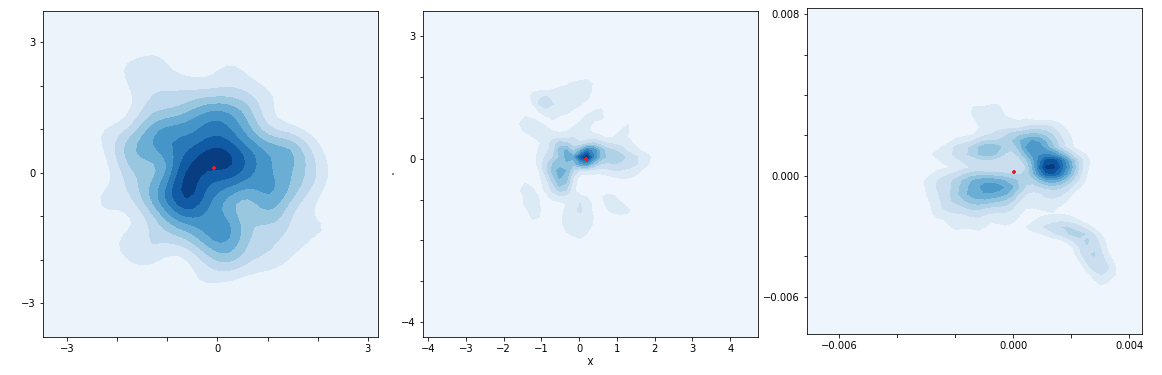}
\caption{Comparing the prior Gaussian  distribution $p(z)$, MMD $q_{\phi}(z)$ and KL $q_{\phi}(z)$. The \textcolor{red}{red dots} represent $(0,0)$. It clearly demonstrates that with KL $q_{\phi}(z)$, the distribution matches the prior Gaussian distribution $p(z)$ poorly, while with MMD $q_{\phi}(z)$ matches significantly better.}
\label{fig:mmd}
\end{figure}

To make this more intuitive, we conduct experiment on MNIST dataset \citep{lecun1998gradient}. For visualization, we make the latent code two dimensions. It can be seen in Figure \ref{fig:mmd} that with KL $q_{\phi}(z)$, the distribution matches the prior Gaussian distribution $p(z)$ poorly, while with MMD $q_{\phi}(z)$ matches significantly better. And results in Table \ref{tab: feature visualization} also demonstrate that MMD helps better reconstruction than KL.

\begin{table}[t]
  \centering
 \caption{Feature visualization.}
\label{tab: feature visualization}
{
\begin{tabular}{ccc}
\toprule
Method                  &Reconstruction loss      & Log likelihood  \\
\midrule
KL                          &	0.04367 		                &82.75                  \\
MMD                     & \textbf{0.03605}	  	    &\textbf{80.76 }                 \\

\bottomrule
\end{tabular}
}
\end{table}

\section{Conclusion}
 In this paper, we propose a unified framework for learning to generate diverse outputs with unpaired training data and allow simultaneous training of multiple datasets with different domains by a single network. Furthermore, we also investigate how to better extract domain supervision information so as to better utilize domain information and explicitly constrain the disentanglement. Qualitative and quantitative experiments on different datasets show that the proposed method outperforms the state-of-the-art methods.

%\begin{acknowledgements}
%If you'd like to thank anyone, place your comments here
%and remove the percent signs.
%\end{acknowledgements}

% BibTeX users please use one of
\bibliographystyle{spbasic}      % basic style, author-year citations
\bibliography{reference}   % name your BibTeX data base

\section{Appendix}
\begin{figure*}
  \includegraphics[width=1.0\textwidth]{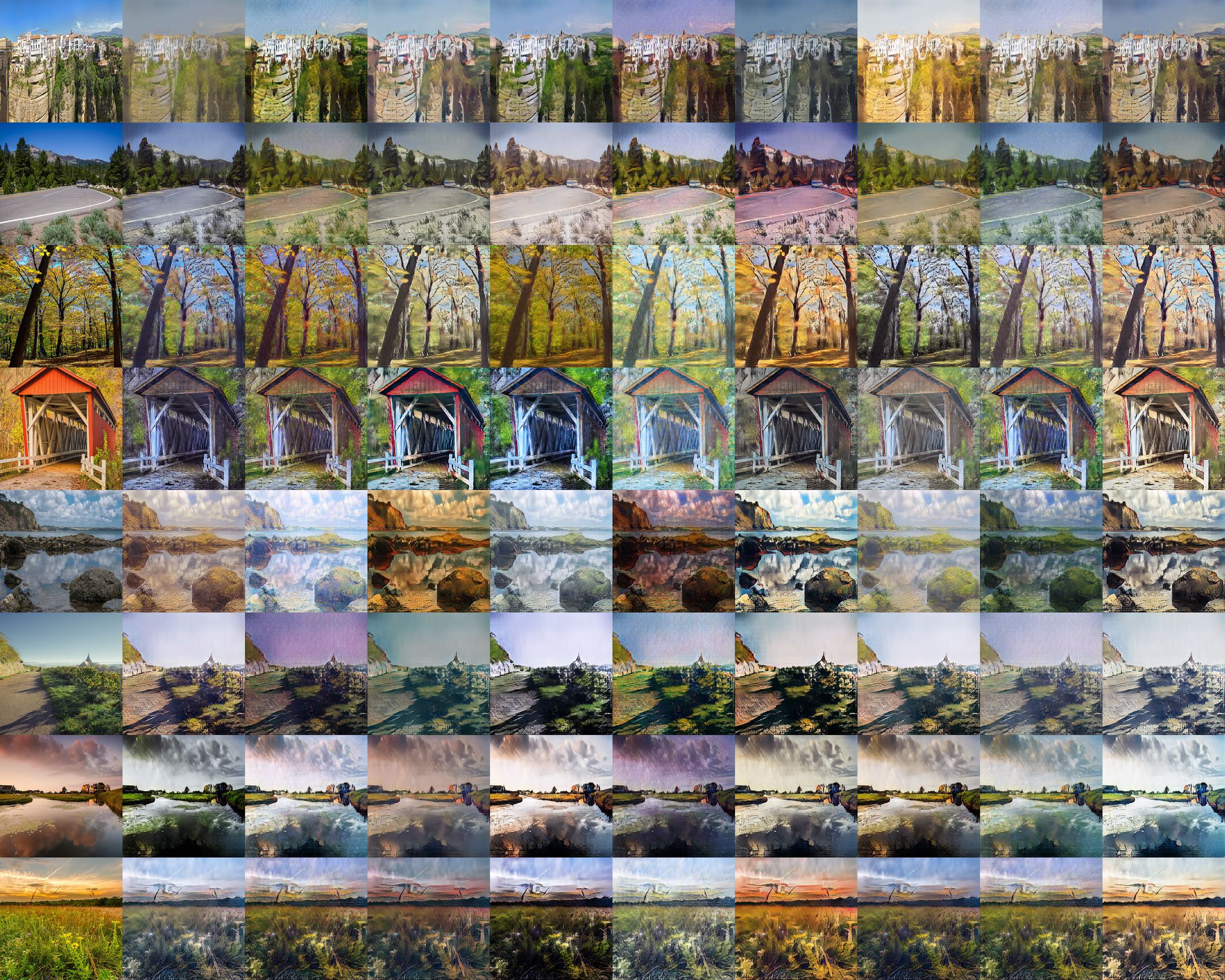}
\caption{Art result. Better look by zooming in.}
\label{fig:art_d4_result}
\end{figure*}

\begin{figure*}
  \includegraphics[width=1.0\textwidth]{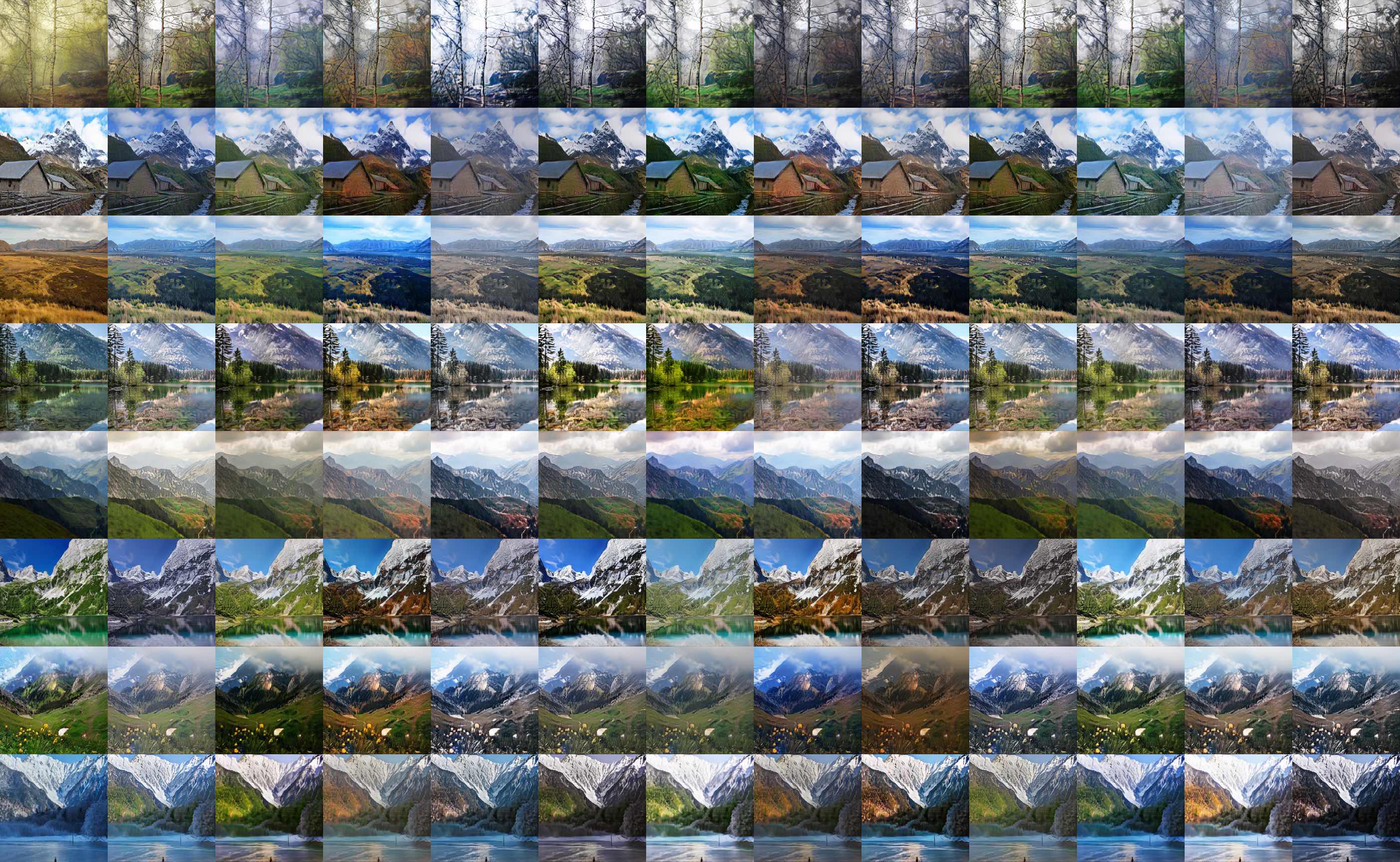}
\caption{Season result. Better look by zooming in.}
\label{fig:season_d4_result}
\end{figure*}

\begin{figure*}
  \includegraphics[width=1.0\textwidth]{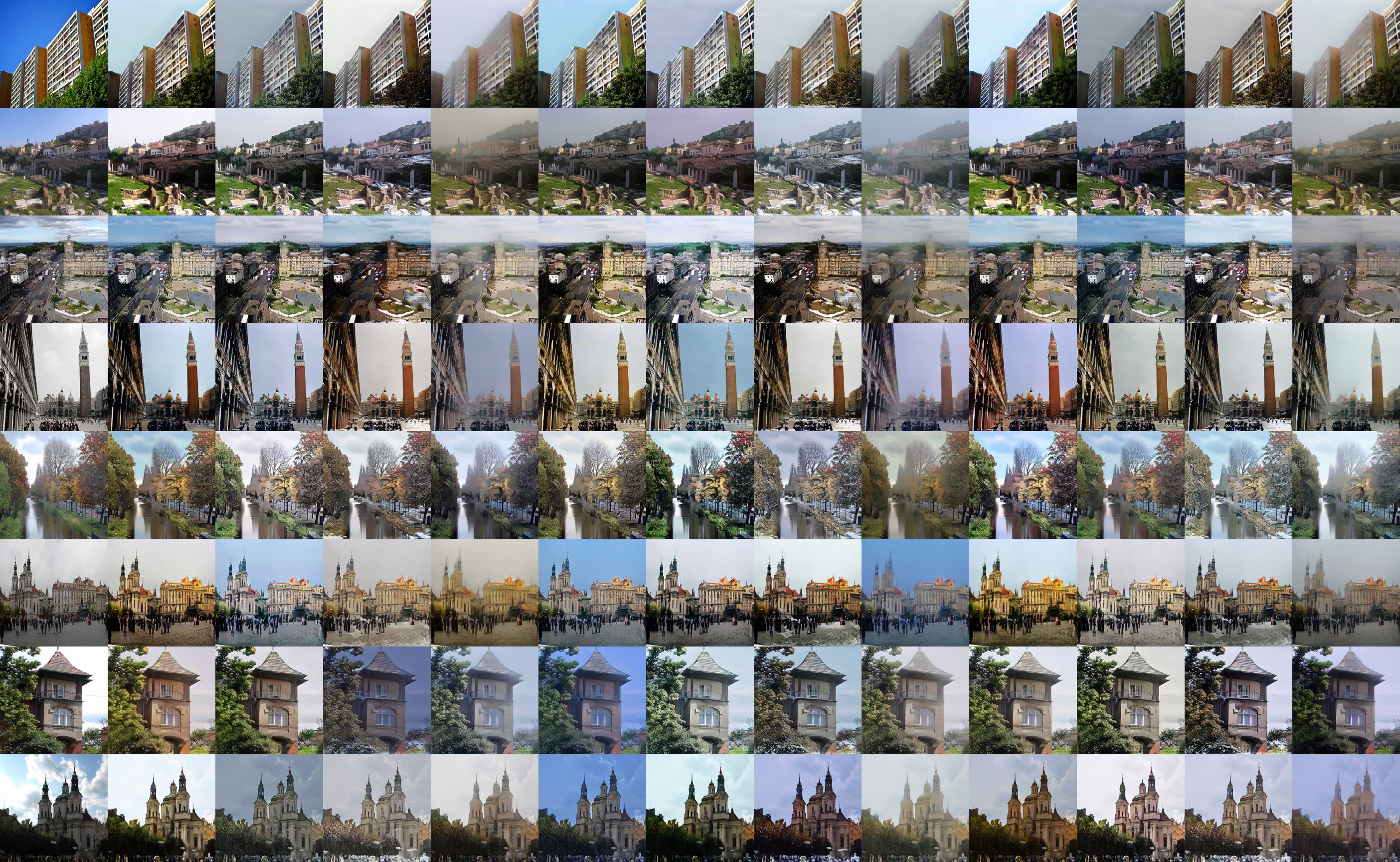}
\caption{Weather result. Better look by zooming in.}
\label{fig:weather_d4_result}
\end{figure*}

In this appendix, we show some additional cross-domain translation results of art in Figure \ref{fig:art_d4_result}, season in Figure \ref{fig:season_d4_result} and weather in Figure \ref{fig:weather_d4_result}.

\end{document}